\documentclass{ecai} 

\usepackage{latexsym}
\usepackage{amssymb}
\usepackage{amsmath}
\usepackage{amsthm}
\usepackage{booktabs}
\usepackage{enumitem}
\usepackage{graphicx}
\usepackage{color}

\usepackage{subfigure}
\usepackage{algorithm}
\usepackage{algorithmic} 

\usepackage{multirow} 

\usepackage[commandnameprefix=always]{changes}
\newcommand*{\imgg}[1]{%
    \raisebox{-.02\baselineskip}{%
        \includegraphics[
        height=\baselineskip,
        width=\baselineskip,
        keepaspectratio,
        ]{#1}%
    }%
}

\newtheorem{theorem}{Theorem}
\newtheorem{lemma}[theorem]{Lemma}

\newcommand{\BibTeX}{B\kern-.05em{\sc i\kern-.025em b}\kern-.08em\TeX}

\begin{document}


\begin{frontmatter}


\paperid{123} 


\title{CDSA: Conservative Denoising Score-based Algorithm for Offline Reinforcement Learning}


\author[A]{\fnms{Zeyuan}~\snm{Liu}\thanks{Equal contribution.}}
\author[A]{\fnms{Kai}~\snm{Yang}\footnotemark[*]}
\author[A]{\fnms{Xiu}~\snm{Li}\thanks{Corresponding Author. Email: li.xiu@sz.tsinghua.edu.cn}} 

\address[A]{Tsinghua Shenzhen International Graduate School, Tsinghua University}



\begin{abstract}
Distribution shift is a major obstacle in offline reinforcement learning, which necessitates minimizing the discrepancy between the learned policy and the behavior policy to avoid overestimating rare or unseen actions. 
Previous conservative offline RL algorithms struggle to generalize to unseen actions, despite their success in learning good in-distribution policy. In contrast, we propose to use the gradient fields of the dataset density generated from a pre-trained offline RL algorithm to adjust the original actions. We decouple the conservatism constraints from the policy, thus can benefit wide offline RL algorithms. As a consequence, we propose the Conservative Denoising Score-based Algorithm (CDSA) which utilizes the denoising score-based model to model the gradient of the dataset density, rather than the dataset density itself, and facilitates a more accurate and efficient method to adjust the action generated by the pre-trained policy in a deterministic and continuous MDP environment. In experiments, we show that our approach significantly improves the performance of baseline algorithms in D4RL datasets, and demonstrate the generalizability and plug-and-play capability of our model across different pre-trained offline RL policy in different tasks. We also validate that the agent exhibits greater risk aversion after employing our method while showcasing its ability to generalize effectively across diverse tasks.
\end{abstract}

\end{frontmatter}


\section{Introduction}

Reinforcement learning (RL) algorithms have been demonstrated to be successful on a range of challenging tasks, from games \cite{mnih2013playing, silver2016mastering} to robotic control \cite{schulman2015trust}. As one of its branches, offline RL only uses a fixed training dataset during training, which aims at finding effective policies while avoiding online interactions with the real environment and a range of related issues \cite{lange2012batch,fujimoto2019off}. However, due to the data-dependent nature of offline RL algorithms, this pattern makes distribution shift a major obstacle to the effectiveness of the algorithm \cite{fujimoto2019off,lyu2022mildly}. Offline RL algorithms are prone to producing inaccurate predictions and catastrophic action commands when queried outside of the distribution of the training data, which leads to catastrophic outcomes. To strike a suitable trade-off between learning an improved policy and minimizing the divergence from the behavior policy, aiming to avoid errors due to distribution shift, previous work has provided various perspectives, including constraining the system in the training dataset distribution \cite{kumar2019stabilizing,fujimoto2019off,lyu2023state}, or developing a distributional critic to leverage risk-averse measures \cite{ma2021conservative,urpi2021risk}.

However, most previous conservative offline RL algorithms failed to fully disentangle the knowledge related to conservatism from the algorithm's training process. This knowledge is typically incorporated into functions such as the final policy or critics, rendering it inseparable from other components. Consequently, even if the training dataset remains unchanged, various algorithms are unable to directly exchange their conservatism-related knowledge, particularly if this knowledge is deemed solely dependent on the distribution of the training dataset. To tackle this issue, we explore the possibility of learning conservatism-related knowledge exclusively from the training dataset to obtain a plug-and-play decision adjuster. One intuitive approach is to leverage the density distribution of each dataset to guide the agent towards states located in areas of high density as much as possible. This can be achieved by adjusting the actions within the dataset to steer transitions towards states with higher density. Essentially, this modification makes the executed actions safer and more conservative. This approach enables us to utilize any offline RL algorithm as the baseline algorithm. By ensuring the algorithm uses the same training dataset, we can effectively harness the acquired plug-and-play conservatism-related knowledge to enhance the algorithm's performance.

To obtain the distribution of the dataset, various methods commonly utilize the approach of reconstructing the density within the training dataset \cite{kumar2019stabilizing,fujimoto2019off,richter2017safe,gelada2019off,mcallister2019robustness}. Previous studies in this domain have utilized density models to restrict or regularize the controller, preventing the agent from selecting actions or exploring states with low likelihood in the dataset. In contrast, our approach does not impose constraints but rather guides the agent. Specifically, during each step of the decision-making process, we generate multiple action components to supplement the original action produced by a pre-trained offline RL algorithm. This encourages the agent to select actions that are more prevalent in the dataset and transition to states with higher likelihoods, mitigating inaccuracies caused by distribution shifts. Additionally, if the agent encounters a low-likelihood state, our method provides guidance to navigate away from regions lacking sufficient support in the training data.

\begin{figure*}[t]
    \centering
    \subfigure[]{
    \begin{minipage}[t]{0.9\linewidth}
    \centering
    \includegraphics[height=60pt,width=300pt]{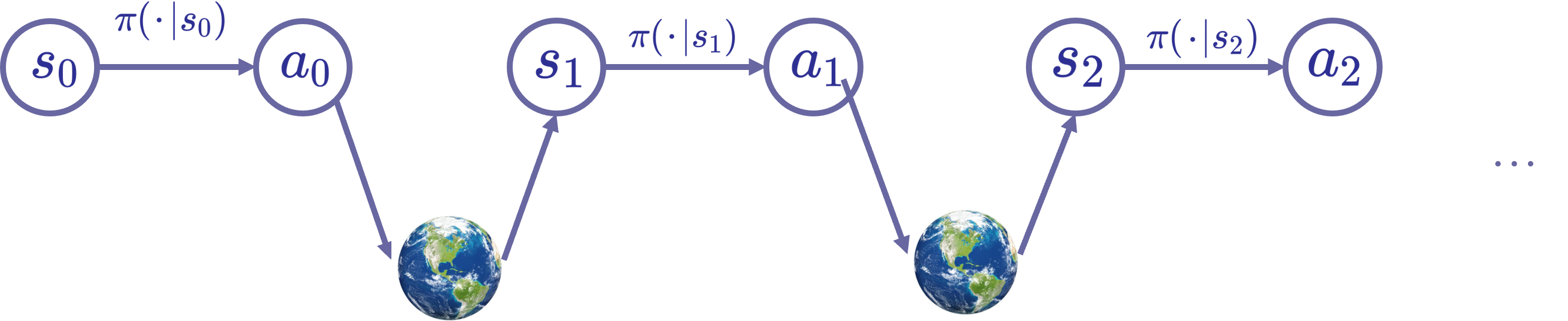}
    \end{minipage}
    }
    \subfigure[]{
    \begin{minipage}[t]{0.9\linewidth}
    \centering
    \includegraphics[height=83pt,width=350pt]{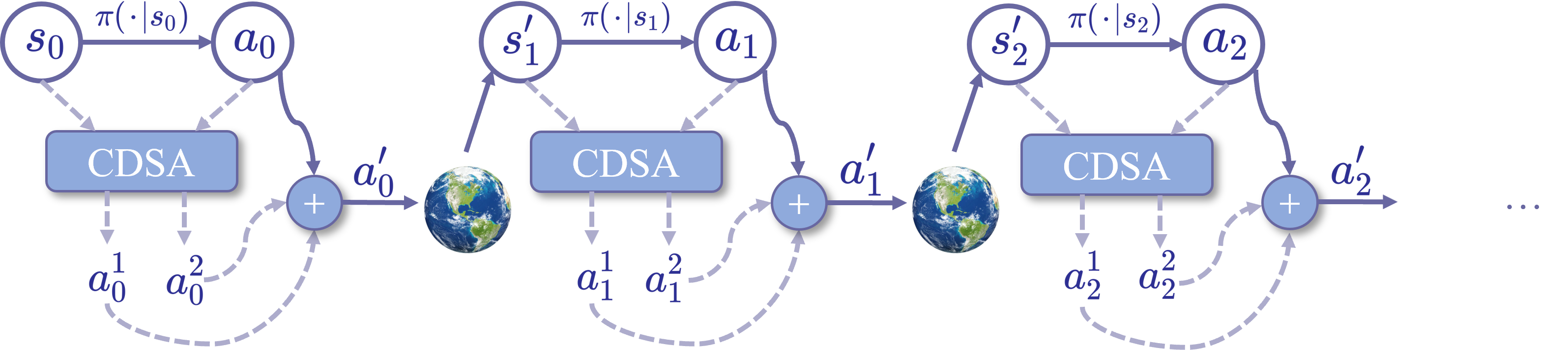}
    \end{minipage}
    }
    \caption{The Conservative Denoising Score-based Algorithm (CDSA) leverages conservatism-related knowledge to enhance the performance of offline RL algorithms. As depicted in (a), the original RL algorithm generates actions based on the current state to interact with the environment. To address the distribution shift problem, we propose to generate auxiliary actions based on the current action-state pair, guiding the entire trajectory towards high-density regions of the training dataset. This is illustrated in (b), where CDSA generates two action components, utilizing conservatism-related  knowledge acquired from the training dataset, to be added to the action generated from a pre-trained policy $\pi$.}
    \label{idea}
\end{figure*}
\setcounter{figure}{0}

Combining these altogether, we propose the Conservative Denoising Score-based Algorithm (CDSA), which does not intervene in the training process of the original offline RL algorithm. CDSA adjusts the generated actions during testing while avoiding excessive interference with the original decisions. It also proactively guides the trajectory into regions of higher density in the training dataset as shown in Figure \ref{idea}. Our idea is similar to the approach used in Lyapunov Density Models (LDM) \cite{kang2022lyapunov}, which employs strong theoretical constraints and model-based planning methods to keep the agent confined to high-density regions of the training dataset. However, in the LDM approach, the density of early termination points must be manually labeled to maintain low density for specific experiments (as outlined in section D.2 of \cite{kang2022lyapunov}). This introduces subjective human knowledge and imposes stringent constraints that restrict the algorithm's versatility. In contrast, our methodology operates without such limitations. Moreover, while LDM exclusively deals with scenarios within the training data distribution, our approach also incorporates guiding the agent outside the training data distribution. Furthermore, it is worth noting that while LDM trains a Dynamics Model and performs MPC, recurrent use of the dynamics model may result in accumulating inference errors. To address the adverse effects stemming from network uncertainty, we adopt a strategy where, in each step, we employ the inverse dynamics model for inference only once. This practice effectively minimizes errors.

Our experiments demonstrate that CDSA directly enhances the performance of Offline RL baseline algorithms and can be applied to different algorithms without any fine-tuning or new conservatism-related knowledge learning. The baseline algorithms, IQL \cite{kostrikov2021offline} and POR \cite{xu2022policy}, exhibit improvements of 12.7\% and 5.2\% respectively when enhanced with CDSA on d4rl dataset. We also conducted supplementary experiments, including the ablation study on the effect of auxiliary actions in Appendix \ref{section:C}.

\section{Related Work}

\subsection{Offline RL} Offline (or batch) RL  aims to learn a policy using a fixed dataset collected by some unknown behavior policies \cite{lange2012batch,levine2020offline}. The critical challenge in offline RL is the distribution shift \cite{fujimoto2019off,kumar2019stabilizing}, where the agent overestimates and prefers the out-of-distribution (OOD) actions, with the result that it performs poorly. Currently, there are various solutions to this problem involving constraining the learned policy to be closer to the original behavior policy \cite{fujimoto2021minimalist,kumar2019stabilizing,wang2022diffusion}, regularizing the critic learning to be more pessimistic with OOD actions \cite{kostrikov2021offline,kumar2020conservative,yang2024exploration}, adopting model-based approaches \cite{yu2020mopo,diehl2021umbrella,lyu2022doublecheck,Zhang2023UncertaintydrivenTT}, leveraging uncertainty measurement \cite{bai2022pessimistic,wu2021uncertainty,an2021uncertainty}, importance sampling \cite{sutton2016emphatic,liu2019off,gelada2019off}, etc. In particular, some works learn the density model of the training data to help constrain the agent in distribution \cite{richter2017safe,mcallister2019robustness,fujimoto2019off,kumar2019stabilizing}. We mainly draw lessons from these works and propose to learn a gradient field, which can solve the bootstrapping error problem more flexibly.

\subsection{Risk-averse RL} In risk-averse reinforcement learning, the goal is to optimize some risk measure of the returns. The most common risk-averse measures are the Value-at-Risk (VaR) and Conditional Value-at-Risk (CVaR) \cite{rockafellar2002conditional}, which use quantiles to parameterize the policy return distribution. There are also other measures such as the Wang measure \cite{wang1996premium}, the mean-variance criteria \cite{namkoong2017variance}, and the cumulative probability weighting (CPW) metric \cite{tversky1992advances}. However, our goal is not to optimize these criteria but to use them as an evaluation indicator of the conservatism of algorithms. Our work mainly focuses on VaR and uses it as a measure to evaluate the risk-averse ability of our algorithm.

\subsection{Scored-based model} Recently, score-based generative models \cite{song2019generative,song2020score,vahdat2021score} have received much attention. The main idea of the score-based model is that the probability distribution of real data is represented by score \cite{liu2016kernelized}, a vector field, which points to the direction where the data is most likely to increase. Leveraging the learned score function as a prior, we can perform the Langevin Markov chain Monte Carlo sampling \cite{welling2011bayesian} to generate the desired data from random noise \cite{vincent2011connection}. The score-based model has been successful at generating data from various modalities such as images \cite{ho2020denoising,song2020score}, audio \cite{kong2020diffwave} and graphs \cite{niu2020permutation}. Some methods introduce score-based models into RL \cite{wu2022targf}, which try to use the score-based model to solve the problem of object rearrangement. It only considers how to improve the probability of the states in the original distribution and must participate in the training process of the baseline algorithm. In our work, we use the loss from denoising score matching \cite{vincent2011connection}, which is more concise than the original score-based method. 


\section{Preliminaries}

We consider the Markov Decision Process (MDP) setting \cite{sutton1998introduction} with continuous states $s\in \mathcal{S}$, continuous actions $a\in \mathcal{A}$, transition probability distribution function $P(\cdot|s,a)$, reward function $r(s,a)$, initial state distribution $\rho$ and discount factor $\gamma$. The aim of RL is to learn a policy $\pi(a|s)$ that maximizes the cumulative discounted returns

\begin{align}
\pi^{*}=\underset{\pi}{\arg \max \ } \mathbb{E}_{\pi}\left[\sum_{t=0}^{\infty} \gamma^{t} r\left(s_{t}, a_{t}\right)\right].
\end{align}

We consider the offline settings in our work. In the offline setting, we only have access to a fixed dataset $D=\{(s,a,r,s')\}$ consisting of trajectories collected by different policies. The agent cannot directly interact with the real environment and will encounter extrapolation errors when visiting OOD states or taking OOD actions. 

The score \cite{liu2016kernelized} of probability density $p_{\text{data}}(\mathbf{x})$ is commonly defined as $\nabla_{\mathbf{x}} \log p_{\text{data}}(\mathbf{x})$. Score matching \cite{hyvarinen2005estimation} is able to estimate $\nabla_{\mathbf{x}} \log p_{\text{data}}(\mathbf{x})$ without training a model to estimate $p_{\text{data}}(\mathbf{x})$. The objective of score matching model minimizes  
\begin{align}
\frac{1}{2} \mathbb{E}_{p_{\text{data}}(\mathbf{x})}\left[\left\|s_{\theta}({\mathbf{x}})-\nabla_{\mathbf{x}} \log p_{\text{data}}(\mathbf{x})\right\|_{2}^{2}\right].
\end{align}
Once the score function is known, the approximate samples for $p_{\text{data}}(\mathbf{x})$ can be generated by Langevin dynamics. The Langevin method recursively computes the following
\begin{align}
\mathbf{x}_t \gets \mathbf{x}_{t-1} + \alpha \nabla_{\mathbf{x}} \log p_{\text{data}}(\mathbf{x}_{t-1}) + \sqrt{2\alpha}z_t,
\end{align}
where $z_t \sim N(0,I)$ and $\alpha$ is a constant. The noise added to the equation is to prevent multiple data points from mapping to the same location within the distribution. When $\alpha$ is small enough and the number of iterations is large enough, the distribution of $\mathbf{x}_t$ is considered to be the same as $p_{\text{data}}(\mathbf{x})$. The $\nabla_{\mathbf{x}} \log p_{\text{data}}(\mathbf{x})$ in this formula is substituted by $s_{\theta}(x)$ since the score network is a good estimation of $\nabla_{\mathbf{x}} \log p_{\text{data}}(\mathbf{x})$.

Suppose states and actions obey an unknown probability data distribution $p_{\text{data}}(s,a)$ in the environment. The dataset consists of i.i.d. samples $\{(s,a)_i\ s\in S,a\in A\}_{i=1}^N$ from $p_{\text{data}}(s,a)$. Suppose we have a pre-trained offline policy $\pi(a|s)$. Our goal is to make the action more conservative, in other words, to increase the probabilities of $p_{\text{data}}(s,a)$.

\section{Methodology}
We use offline datasets that consist of trajectories sampled from any unknown policy to generate gradient fields for action correction. We modify the original action $a_o$ by introducing two action correction terms associated with the action and the state respectively: $a \leftarrow a_o + K_1 * a_1 + K_2 * a_2$, where $a_o$ is the original action sampled from the policy $\pi$, $a_1$ and $a_2$ encourages the agent to favor high likelihood regions, $K_1$ and $K2$ are hyperparameters. The challenge lies in designing $a_1$ and $a_2$ given the unknown grounded distribution $p_{\text{data}}(s, a)$. We draw inspiration from the score matching method to tackle this issue and propose our solution. We do not learn an extra critic or actor network but learn gradient fields to make the agent in or close to distribution.

\subsection{Learning the Density Gradient Fields from Data}

We consider the situation where the offline dataset contains a large fraction of trajectories, and we refer to the probability distribution of state-action pairs in this dataset as $p_{\text{data}}(s,a)$. Our goal is to get our trajectory as close to the distribution $p_{\text{data}}(s,a)$ of the dataset as possible so that the agent can make better decisions. To make the agent in distribution, we need to find directions to increase the log-likelihood of the density for both state and action, and the fastest direction is $\nabla_{(s,a)} \log p_{\text{data}}(s,a)$. What we want to do is to generate auxiliary actions based on the current action-state pair to increase the probability of the entire trajectory within the distribution, which is shown in Figure \ref{idea}. To find an approach to achieve this goal, we refer to the denoising score-matching model \cite{ho2020denoising} and use networks to find these directions.

We adapt the approach of approximating the gradient of points from the denoising score-matching model to offline RL. While this model can easily learn the gradient of points since the coordinates are independent, obtaining the gradient of state-action pairs in RL settings is challenging due to the dependency between states and actions. To address this, we utilize a score-matching model to learn the gradients of actions and states separately.

Firstly, for the gradient of the action, we consider learning the gradient by using a network $g_{\theta}({s,a})$ to fit $\nabla_{a} \log p_{\text {data }}(s,a)$ where $\theta$ is the parameters of this network. To train this network, we adopt the denoising score-matching objective \cite{vincent2011connection}, which guarantees a reasonable estimation of the score. For simplicity, we define $\mathbf{x}=(s,a)$ as current state-action pair and define $\mathbf{\tilde{x}}=(\tilde{s},\tilde{a})$ as the noised state-action pair. The pre-specified noise distribution $q_{\sigma}(\mathbf{\tilde{x}|\mathbf{x}})$ is used to perturb $\mathbf{x}$ and the target of the network is to learn the score of the perturbed target distribution. The loss of the network is

\begin{small}
{
\begin{align}
\mathcal{L}_\theta=
\frac{1}{2} \mathbb{E}_{q_{\sigma}(\tilde{\mathbf{x}} \mid \mathbf{x}) p_{\text {data }}(\mathbf{x})}\left[\left\|g_{\theta}(\tilde{\mathbf{x}})-\nabla_{\tilde{a}} \log q_{\sigma}(\tilde{\mathbf{x}} \mid \mathbf{x})\right\|_{2}^{2}\right].
\end{align}
}
\end{small}

The optimal network satisfies $g_{\theta^{*}}(\mathbf{x})=\nabla_{a} \log q_{\sigma}(\mathbf{x})$ and $\nabla_{a} \log q_{\sigma}(\mathbf{x}) \approx \nabla_{a} \log p_{\text {data }}(\mathbf{x})$. When the loss is small enough, it can be thought that $g_{\theta^{*}}(\mathbf{x})\approx\nabla_{a} \log p_{\text {data }}(\mathbf{x})$. When we choose the pre-specified noise distribution as the normal distribution, the relationship between original data $\mathbf{x}=(s,a)$ and perturbed data $\mathbf{x}=(\tilde{s},\tilde{a})$ is
\begin{align}
\tilde{s}  = s, \tilde{a} \sim N(a, \sigma I).
\end{align}

\setcounter{figure}{1}
\begin{figure}[t]
    \begin{centering}
      \includegraphics[height=135pt,width=225pt]{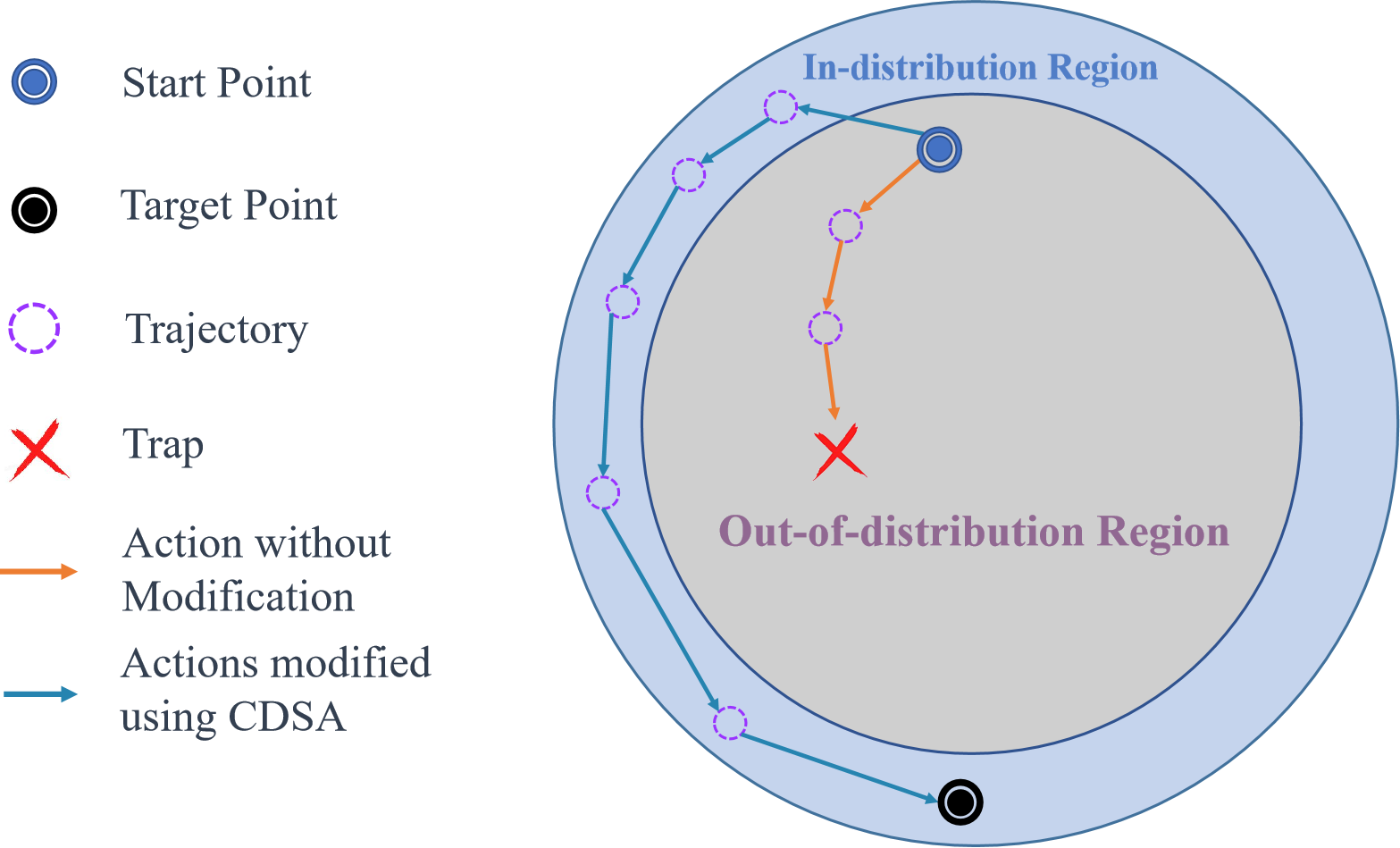}
      \caption{An example comparing CDSA control and common control. CDSA controls more closely to areas of distribution for action decision making.}
    \end{centering}
\end{figure}

Directly optimizing Eq. (4) is difficult because we do not have the direct access to the $\nabla_{\tilde{a}} \log q_\sigma(\tilde{\mathbf{x}} \mid \mathbf{x})$. Thanks to the help of diffusion score-based model, we show in the following lemma that the loss can be rewritten to a simpler form:

\begin{lemma}\label{theorem}
    The loss $\mathcal{L}_\theta$ in Eq. (4) is equivalent to the following loss:
    \begin{align}
    \mathcal{L}_\theta=\frac{1}{2} \mathbb{E}_{q_{\sigma}(\tilde{\mathbf{x}} \mid \mathbf{x}) p_{\text {data }}(\mathbf{x})}\left[\left\|g_{\theta}(\mathbf{x}+\sigma\mathbf{z})+\frac{z}{\sigma}\right\|_{2}^{2}\right],
    \end{align}
    where $\mathbf{z}=(0, z)$ and $z\sim N(0,I)$.
\end{lemma}

The proof of the Lemma 1 can refer to Appendix \ref{section:A.1}. When the loss converges, $g_{\theta}(\mathbf{x})\approx\nabla_{a} \log p_{\text {data }}(\mathbf{x}) $ can be used to modify the action to make it more conservative.

Secondly, for the gradient of the state, we consider learning the gradient by using a network $h_{\varphi }(\mathbf{x})$ to approximate $\nabla_{s}\log p_{\text {data }}(\mathbf{x})$. The approach employed for learning the gradient of states is analogous to that of learning the gradient of actions, wherein the perturbed data can be expressed as:
\begin{align}
\tilde{s} \sim N(s, \sigma I), \tilde{a} = a.
\end{align}

The loss of this network can be expressed as
\begin{align}
\mathcal{L}_\varphi =\frac{1}{2} \mathbb{E}_{q_{\sigma}(\tilde{\mathbf{x}} \mid \mathbf{x}) p_{\text {data }}(\mathbf{x})}\left[\left\|h_{{\varphi }}(\mathbf{x}+\sigma\mathbf{z'})+\frac{z'}{\sigma}\right\|_{2}^{2}\right],
\end{align}
where $\mathbf{z'}=(0, z')$ and $z'\sim N(0,I)$. $h_{\varphi}(\mathbf{x})$ has a reasonable estimation of  $\nabla_{s}\log p_{\text {data }}(\mathbf{x})$ when the loss function converges.

As mentioned earlier, the quickest way to increase $p_{\text {data }}(s,a)$ is through $\nabla_{(s,a)} \log p_{\text{data}}(s,a)$. However, since the current state remains fixed, we cannot change it while we can change the action. Thus, we initially planned to use a forward model $F(s'|s,a)$ to predict the next state based on the current state and action. This simple model-based trick could help us measure the density of the next state in the training dataset and calculate the gradient of the density function. However, the prediction accuracy of this forward model outside the support of the training dataset cannot be guaranteed. Therefore, it becomes challenging for the resulting gradient to effectively guide the changes in action.

To address these issues, we devised a method leveraging an inverse dynamic model. Our aim is to increase the density $p_{\text{data}}$ of agent states in the dataset. An intuitive approach is to utilize gradient descent to identify the direction of increasing $p_{\text{data}}$, thus necessitating the use of gradients of $p_{\text{data}}$ with respect to states $s$ and actions $a$, i.e., $\Delta s = \nabla_s \log p_\text{data} (x), \Delta a = \nabla_a \log p_\text{data}(x)$. In this way, we can adjust the sampeld action $a$ and encourage the agent to come closer to $s+\Delta s$. In our experiments, we found that employing $\Delta{s}=\nabla_{s} \log p_{\text{data}}(\mathbf{x})$ and integrating an inverse dynamic model $I_{\phi}(s,\tilde{s})$ to generate an action as an auxiliary action component can yield more stable and effective results. This method demonstrates greater reliability compared to using $\Delta{s'}=\nabla_{s'} \log p_{\text{data}}( F(\mathbf{x}), \cdot)$, as it relies on only one model-based prediction network, as opposed to the original concept. The efficacy of these auxiliary action components is substantiated through experiments.

The loss of the inverse dynamic network to learn this knowledge, which is learned by imitation learning \cite{ho2016generative,abbeel2004apprenticeship}, is:
\begin{align}
\mathcal{L}_\phi=\mathbb{E}_{s, a, \tilde{s} \sim \mathcal{D}}\left\|I_\phi(s, \tilde{s})-a\right\|_2^2,\label{hahaha}
\end{align}
where $I_\phi(s, \tilde{s})$ is the inverse dynamic network that learns action from state $s$ to state $\tilde{s}$. When the inverse dynamic network is well-trained, we can input the original state and noised state into this network and get the action with good estimation to change $\Delta{s}$.

\begin{algorithm}
\caption{Training CDSA from offline data}
\label{ALGORITHM1}
\begin{algorithmic}[1]
\STATE \textbf{Input:} Dataset $D$, iterations $T$, learning rate $\eta_\theta, \eta_{\varphi}, \eta_\phi$
\STATE Initialize parameters $\theta, \varphi, \phi$
\FOR{$t=1,2,..T$}
    \STATE Sample $\left(s_t, a, s_{t+1}\right) \sim D$
    \STATE Sample $z, z^{\prime} \sim N(0, I)$
    \STATE Calculate $\mathcal{L}_\theta$ by Eq. (6); update $\theta \leftarrow \theta-\eta_\theta \nabla \mathcal{L}_\theta$
    \STATE Calculate $\mathcal{L}_{\varphi}$ by Eq. (8); update $\varphi \leftarrow \varphi-\eta_{\varphi} \nabla \mathcal{L}_{\varphi}$
    \STATE Calculate $\mathcal{L}_\phi$ by Eq. (9); update $\phi \leftarrow \phi-\eta_\phi \nabla \mathcal{L}_\phi$
\ENDFOR
\STATE \textbf{Output:} networks $g_\theta(s, a), h_{\varphi}(s, a), F_\phi\left(s, s^{\prime}\right)$
\end{algorithmic}
\end{algorithm}

\subsection{
Control with CDSA
}

With the essential knowledge covered, we can now utilize the gradient fields. During the sampling process, we obtain the current state $s$ and an original action $a_{\text{o}}$ from a baseline algorithm such as CQL, IQL, and others. From the gradient field, we extract $\Delta{s}$ and $\Delta{a}$, allowing us to compute $a_1=\Delta{a}\approx g_{\theta}(s,a_o)$ and $a_2=I_{\phi}(s,s+\Delta{s}) \approx I_{\phi}(s,s+h_{\varphi}(s,a_o))$. Here, $\Delta{a}$ and $\Delta{s}$ are obtained from the score model, representing the directions in action and state space that maximize dataset density at the current state-action pair. We directly set $a_1$ as $\Delta{a}$ and then feed $\Delta{s}$ into the Inverse dynamics model $I_{\phi}(\cdot, \cdot)$ to obtain $a_2$. $a_1$ and $a_2$ represent the action correction components obtained from both action and state perspectives. $a_1$ We add these two items to the original action $a_{o}$ linearly, which can be written as
\begin{align}
a = a_{\text{o}} + \delta a, \label{CDSA}
\end{align}

where $\delta a = K_1 g_{\theta}(s,a_o) + K_2 I_{\phi}(s,s+h_{\varphi})$ and $K_1,K_2$ are two hyperparameters. 
However, this auxiliary action can only temporarily and quickly increase the probability of the current state-action pair within the distribution, without considering future situations. Therefore, drawing on the idea of generative model, After using the model to obtain the corrected action $a$, we repeatedly put $(s,a)$ into the model to generate a new $a$, and constantly improve the probability of $(s,a)$ in the distribution.

\begin{figure*}[t]
    \centering
    \subfigure[environment]{
    \begin{minipage}[t]{0.31\linewidth}
    \centering
    \includegraphics[height=140pt,width=150pt]{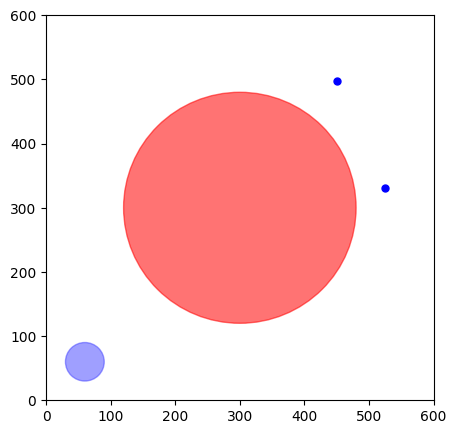}
    \end{minipage}
    }
    \subfigure[search]{
    \begin{minipage}[t]{0.31\linewidth}
    \centering
    \includegraphics[height=140pt,width=150pt]{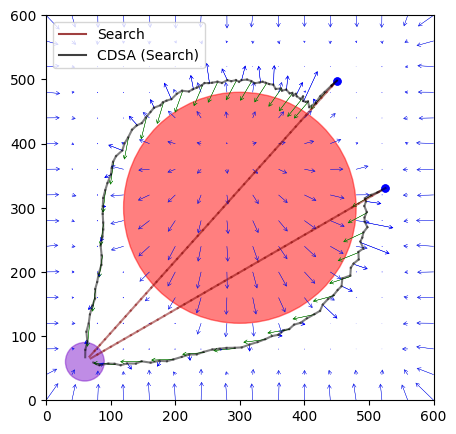}
    \end{minipage}
    }
    \subfigure[CQL]{
    \begin{minipage}[t]{0.31\linewidth}
    \centering
    \includegraphics[height=140pt,width=150pt]{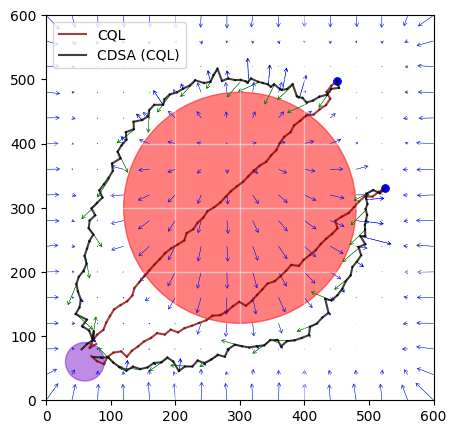}
    \end{minipage}
    }
    \caption{Experiments in the Risky PointMass environment are depicted in (a), where the red circle represents the risky zone, leading to negative rewards if occupied. The agent begins at the blue point, targeting the purple circle. In (b) and (c), we employ simple shortest path finding and CQL as baseline algorithms, demonstrating the corrective impact of our method. CDSA learns from an offline dataset generated by a pretrained CODAC agent, following the identical training procedure outlined in its official repository \cite{ma2021conservative}. Maroon trajectories illustrate baseline algorithm results, while black trajectories depict agents equipped with CDSA. Our method displays two sets of direction arrows: green indicating baseline algorithm directions, and blue indicating conservative auxiliary action directions from CDSA. The arrow length signifies action magnitude. With CDSA modifications, the agent effectively avoids the risky region.}
    \label{FIGURE1}
\end{figure*}

\begin{algorithm}[h]
  \caption{Control with CDSA}
  \label{ALGORITHM2}
  \begin{algorithmic}[1]
    \STATE Input: Environment $E$, pre-trained policy $\pi(a \mid s)$, \\
$\quad$ networks $g_\theta(s, a), h_{\varphi}(s, a), I_\phi\left(s, s^{\prime}\right)$, hyperparameters \\
$\quad K_1, K_2$ \\
    \STATE Get initial State $s$ from $E$, set $d$ as False
    \WHILE{not $d$}
      \STATE Sample original action $a_o$ from $\pi(a \mid s)$
      \STATE Get safety action component $a_1=g_\theta(s, a)$ \\
      \STATE Get safety action component $a_2=I_\phi\left(s, s+h_{\varphi}(s)\right)$ \\
      \STATE  $a \leftarrow a_o + K_1 * a_1 + K_2 * a_2$
      \STATE Roll out $a$ and get $\left(s^{\prime}, r, d\right)$ \\
    \ENDWHILE{}
    \STATE Set $s \leftarrow s^{\prime}$ \\
  \end{algorithmic}
\end{algorithm}

The complete algorithm is presented in Algorithm \ref{ALGORITHM1} and \ref{ALGORITHM2}. Algorithm \ref{ALGORITHM1} trains CDSA using the input dataset and generates three networks. One network serves as the inverse dynamic model, while the other two networks capture information about the gradient fields. In Algorithm \ref{ALGORITHM2}, these networks are utilized to adjust the agent's actions in the given environment. The pre-trained policy used in Algorithm \ref{ALGORITHM2} can be obtained from any RL baseline algorithm, and $K_1$ and $K_2$ are hyperparameters that control the scopes of auxiliary action components.

\begin{table*}[t]
\caption{Average normalized scores of algorithms. We chose 7 popular offline RL algorithms to evaluate the effectiveness of our algorithm. The scores are taken over the final 20 evaluations for MuJoCo and 100 evaluations for AntMaze. CDSA (IQL) and CDSA (POR) achieved the highest scores in 12 out of 15 tasks. The abbreviations in the table correspond to the following meanings: r = random, m = medium, e = expert, u = umaze, l = large, p = play, d = diverse.}
\centering
\large
\setlength{\tabcolsep}{0.5mm}{
    \resizebox{0.75\linewidth}{33mm}{
\begin{tabular}{l c c c c c c c c c c }
\toprule
 \textbf{Dataset} & \textbf{One-step} & \textbf{10\%BC} & \textbf{TD3+BC} & \textbf{CQL}&\textbf{CODAC} & \textbf{IQL} & \textbf{POR} & \textbf{CDSA (IQL)} & \textbf{CDSA (POR)} \\
\midrule
  hopper-r  & 5.2&4.2 &  8.5  & 7.9& 11.0& 10.8&12.5 & 30.9$\pm$0.19 &\textbf{31.9}$\pm$0.14       \\
  hopper-m  & 59.6&56.9& 59.3 &53.0 &70.8 &62.1&89.4 & 65.4 $\pm$5.38& \textbf{90.6}$\pm$10.47   \\
 hopper-m-e &103.3&110.9&  98.0 & 105.6 &\textbf{112.0} &109.5&104.0& \textbf{112.0}$\pm$1.54  &  106.5$\pm$3.21\\
 \hline
 halfcheetah-r &3.7&5.4 & 11.0    & 17.5 &\textbf{34.6} &16.8&17.2& 17.0$\pm$0.21 &17.5$\pm$0.91 \\
 halfcheetah-m &48.4&42.5 &  48.3   & 47.0 & 46.3&48.5&48.1& \textbf{49.1} $\pm$0.38 &48.1$\pm$1.13 \\
 halfcheetah-m-e &\textbf{93.4}&92.9  &  90.7      & 75.6  & 70.4&79.0&81.6 & 81.1$\pm$1.86&85.0$\pm$1.20     \\
 \hline
 walker2d-r &5.6&6.7 & 1.6 & 5.1& \textbf{18.7}&5.9&7.6 & 7.4$\pm$0.20 & 7.9$\pm$0.17  \\
 walker2d-m &81.8&75.0 &  83.7     &   73.3&82.0 &79.6&82.1& 80.2$\pm$1.61&\textbf{83.8}$\pm$4.12 \\
 walker2d-m-e &113.0&109.0 &  110.1  & 113.8&106.0&107.2&111.6 & 113.9$\pm$0.29 & \textbf{114.4}$\pm$1.63  \\
 \hline
 \textbf{Mujoco Average}&57.1&55.9&56.8&55.4&61.3&57.7&61.6&61.9$\pm$1.3&\textbf{65.1}$\pm$2.6
\\
 \hline
 antmaze-u &64.3&62.8&78.6&74.0 &52.8& 76.4     &  88.4 &89.4$\pm$5.55& \textbf{93.4}$\pm$2.07     \\
 antmaze-u-d &60.7&50.2 &71.4&84.0 &38.4& 63.2   & 80.8 &\textbf{86.6}$\pm$4.03& 83.0$\pm$11.85  \\
 \hline
 antmaze-m-p &0.3&5.4& 10.6&61.2 &0.0&65.4 & 88.2 &72.2$\pm$17.05& \textbf{93.8}$\pm$2.05   \\
 antmaze-m-d &0.0&9.8 &3.0&53.7&0.0& 61.0  & 88.0&72.4 $\pm$17.51& \textbf{89.2}$\pm$4.76 \\
 \hline
 antmaze-l-p &0.0&0.0 & 0.2&15.8&1.4&38.0 & 64.6 &53.4$\pm$6.69& \textbf{69.6}$\pm$12.70     \\
 antmaze-l-d &0.0&0.0 & 0.0&14.9&3.8&35.4 & 67.4 &37.0$\pm$4.06& \textbf{70.4}$\pm$7.23   \\
 \hline
 \textbf{AntMaze Average}&20.9&21.4&27.3&50.6&16.1&56.6&79.6&68.5$\pm$9.1& \textbf{83.23}$\pm$6.8   \\
 \bottomrule

\end{tabular}
    }
}
\label{TABLE1}
\end{table*}

\section{Experiments}

We present empirical evaluations of CDSA in this section. We first conducted experiments on environment Risky PointMass and provided visualized results. We then demonstrate the effectiveness of CDSA in offline D4RL \cite{fu2020d4rl} MuJoCo and AntMaze datasets. We also verify the generalizability of CDSA with different tasks in the same environment. Due to space constraints, we place the ablation study in the appendix.

\subsection{Risky PointMass}
Consider an ant robot with the purpose of fast travelling from a random beginning condition to a purple circle. Using red circles, significant expenses are triggered with a low chance, increasing hazards. To train the agent, we use an offline dataset, which is the replay buffer from a CODAC agent. The result is shown in Figure \ref{FIGURE1}. Because the baseline algorithm incorrectly estimates unfamiliar scenes, it chooses to cross dangerous areas to reach the end point. Using CDSA, we can successfully get the agent out of dangerous areas and into familiar situations to make decisions. 

 \begin{figure*}[h]
    \centering
    \includegraphics[height=300pt,width=525pt]{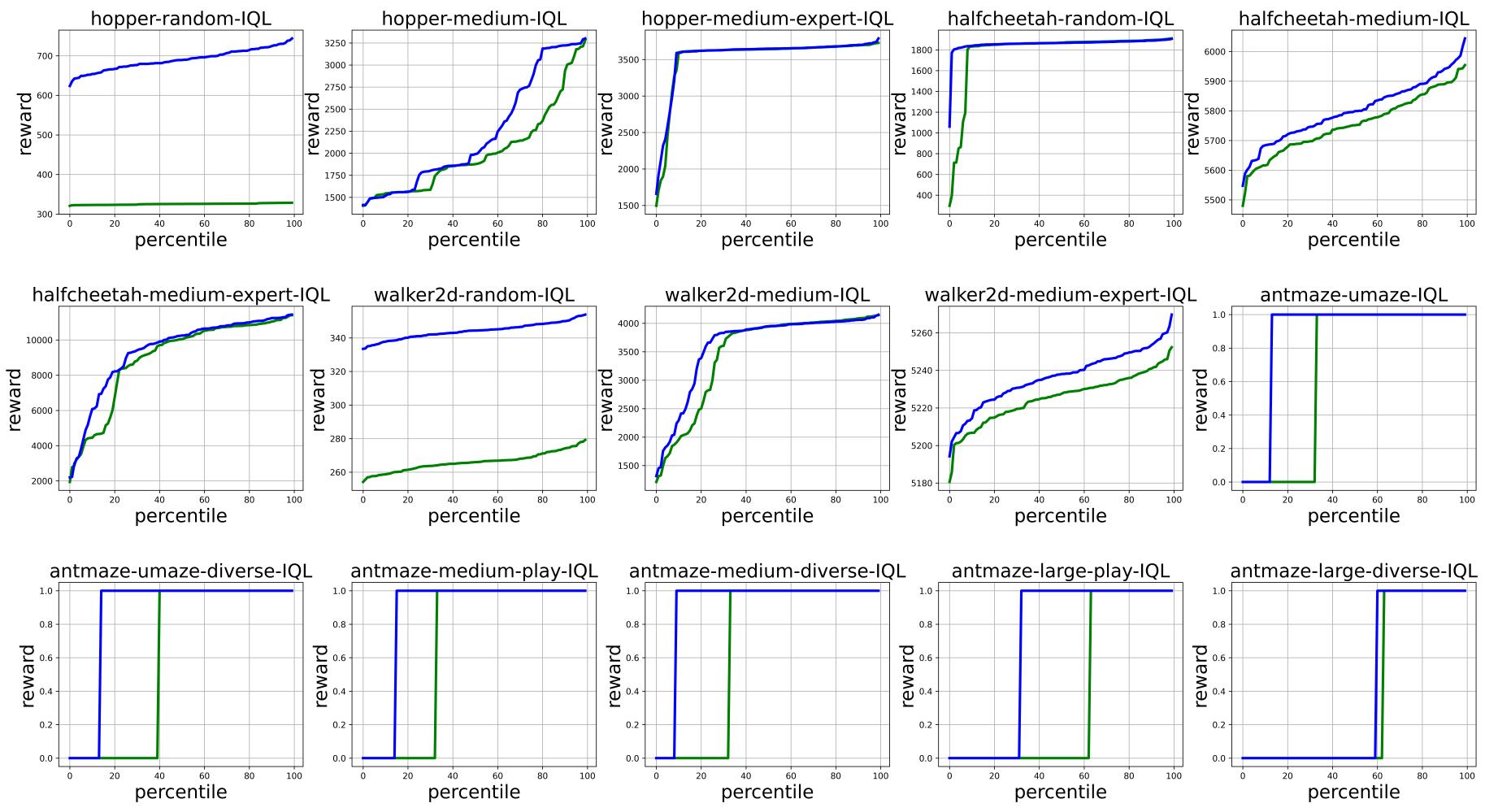}
    \caption{Results of VaR(the $n^{th}$ percentile of cumulative sorted reward). Here only shows the results of CDSA (IQL) (the blue lines) and IQL (the green lines) in the D4RL benchmark. The results of the POR and CDSA (POR) algorithms are shown in Appendix \ref{section:B.1}}
    \label{FIGURE3}
\end{figure*}

\subsection{D4RL offline tasks}
To verify the effectiveness of CDSA in offline scenarios, we evaluate our approach on D4RL MuJoCo and AntMaze datasets. The D4RL MuJoCo benchmark consists of datasets collected by SAC agents that have different performances (random, medium, and expert, etc.) in Hopper, HalfCheetah, and Walker2d environments. The AntMaze environment requires the agent to manipulate a quadruped robot to find the target point in a maze. There are datasets (umaze, medium, and large) divided by the size of the maze. For all datasets we use the "v2" version.
 
In prospect, CDSA should be able to increase the performance of the given policy, hence increasing the final score and VaR (value at risk) in these datasets.

\paragraph{Baselines.} We choose the well-known offline algorithms IQL \cite{kostrikov2021offline} and POR \cite{xu2022policy} as our baseline algorithm. In our experiment, we train models using the IQL and POR methods, each with 5 seeds, and then combine these models with CDSA. We compare the normalized scores with popular RL algorithms, including One-step \cite{brandfonbrener2021offline}, 10\%BC \cite{chen2021decision}, TD3+BC \cite{fujimoto2021minimalist}, CQL \cite{kumar2020conservative}, CODAC \cite{ma2021conservative}, IQL and POR. The experimental details are included in Appendix \ref{section:B.1}.

\begin{figure*}[t]
    \centering
    \subfigure[]{
    \begin{minipage}[t]{0.32\linewidth}
    \centering
    \includegraphics[height=120pt,width=120pt]{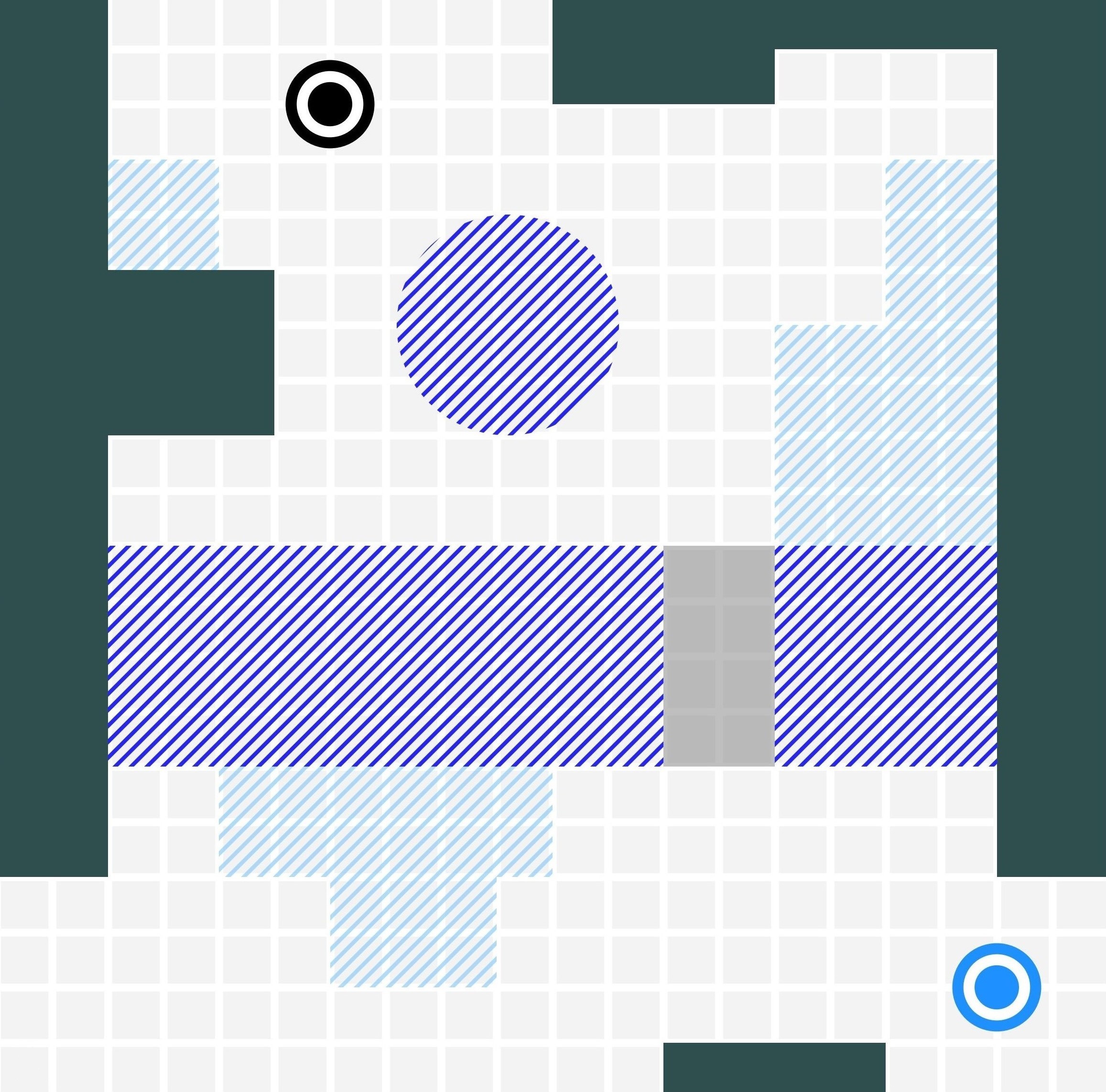}
    \end{minipage}
    }
    \subfigure[]{
    \begin{minipage}[t]{0.32\linewidth}
    \centering
    \includegraphics[height=120pt,width=120pt]{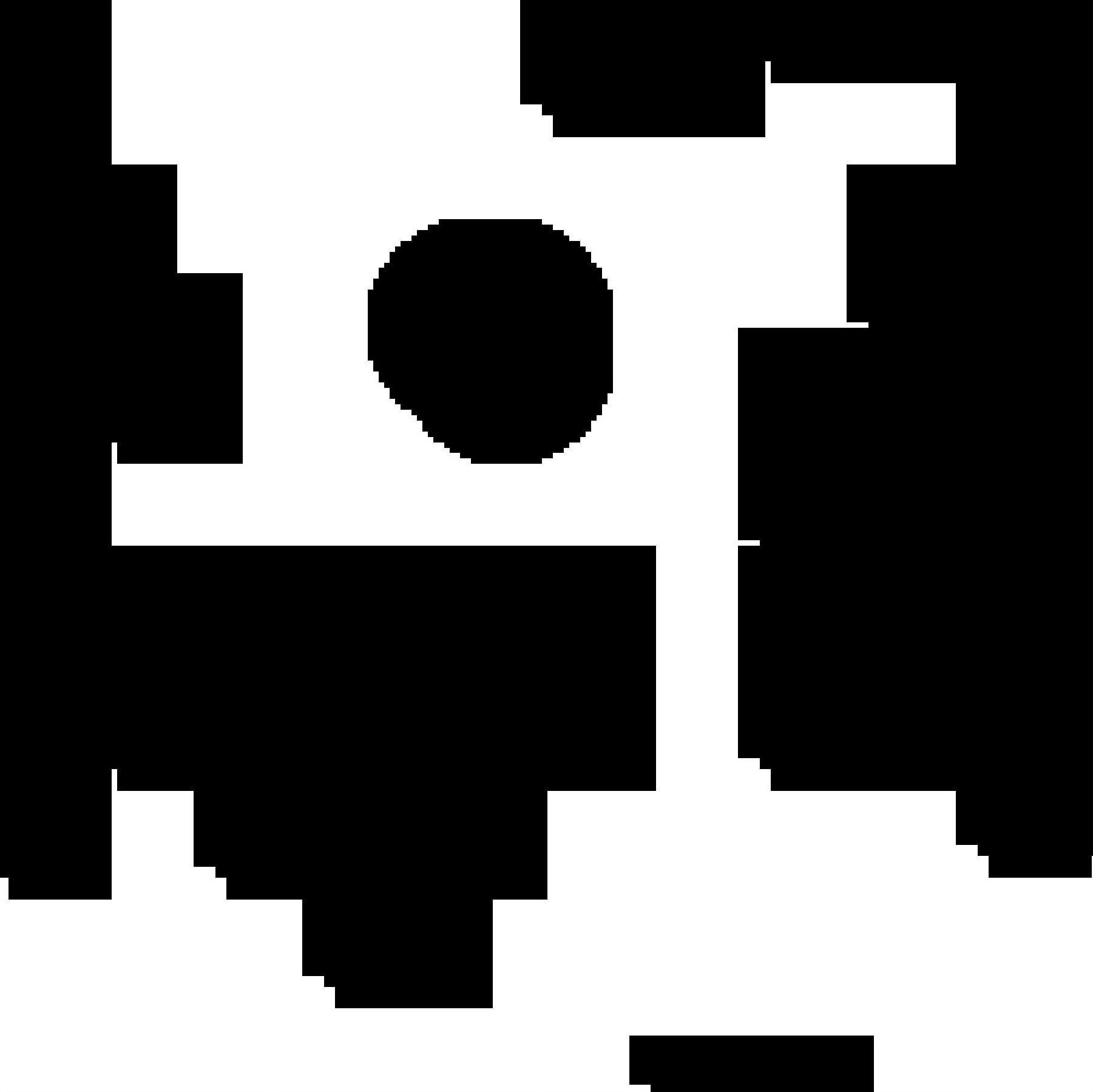}
    \end{minipage}
    }
    \subfigure[]{
    \begin{minipage}[t]{0.32\linewidth}
    \centering
    \includegraphics[height=120pt,width=120pt]{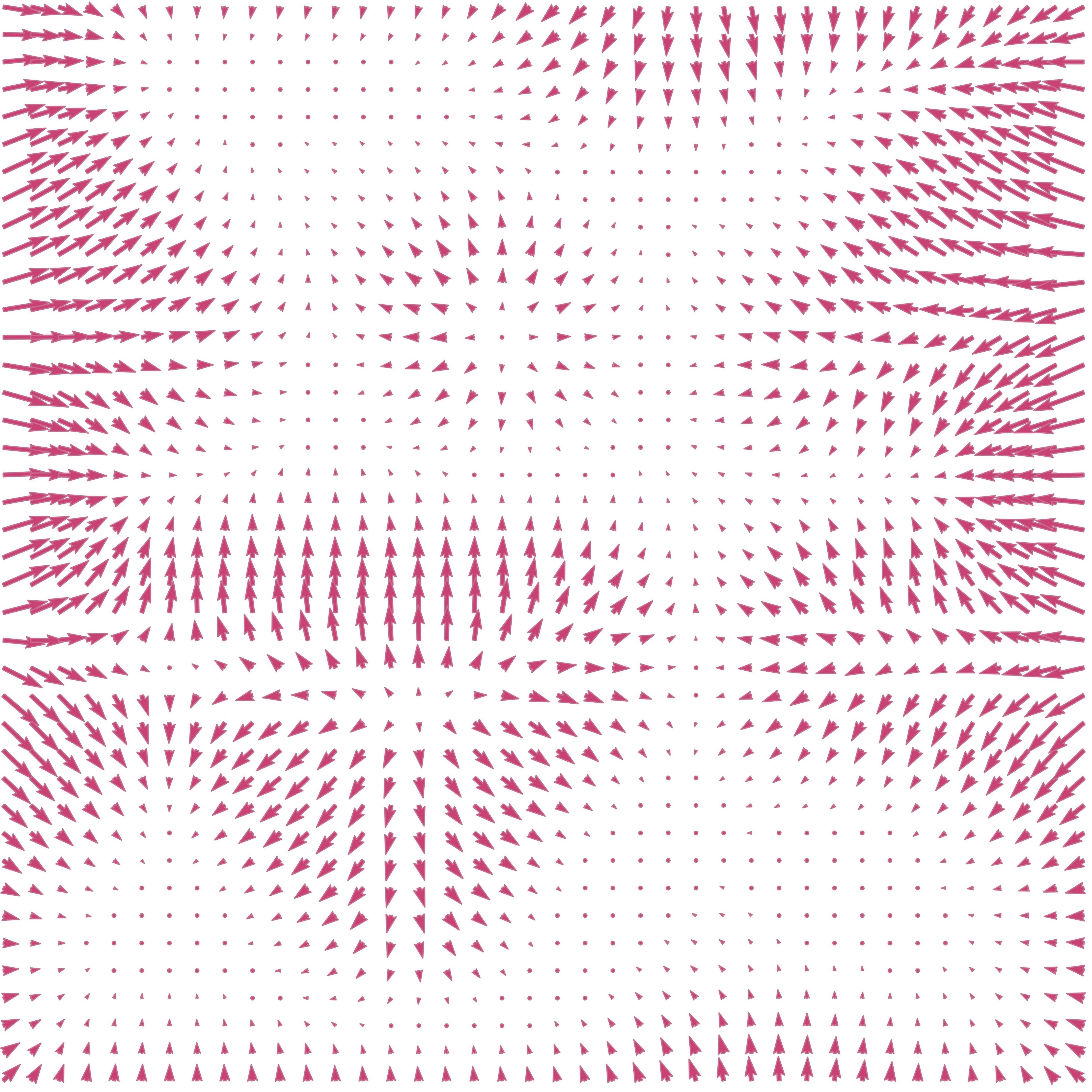}
    \end{minipage}
    }
    \subfigure[]{
    \begin{minipage}[t]{0.32\linewidth}
    \centering
    \includegraphics[height=120pt,width=120pt]{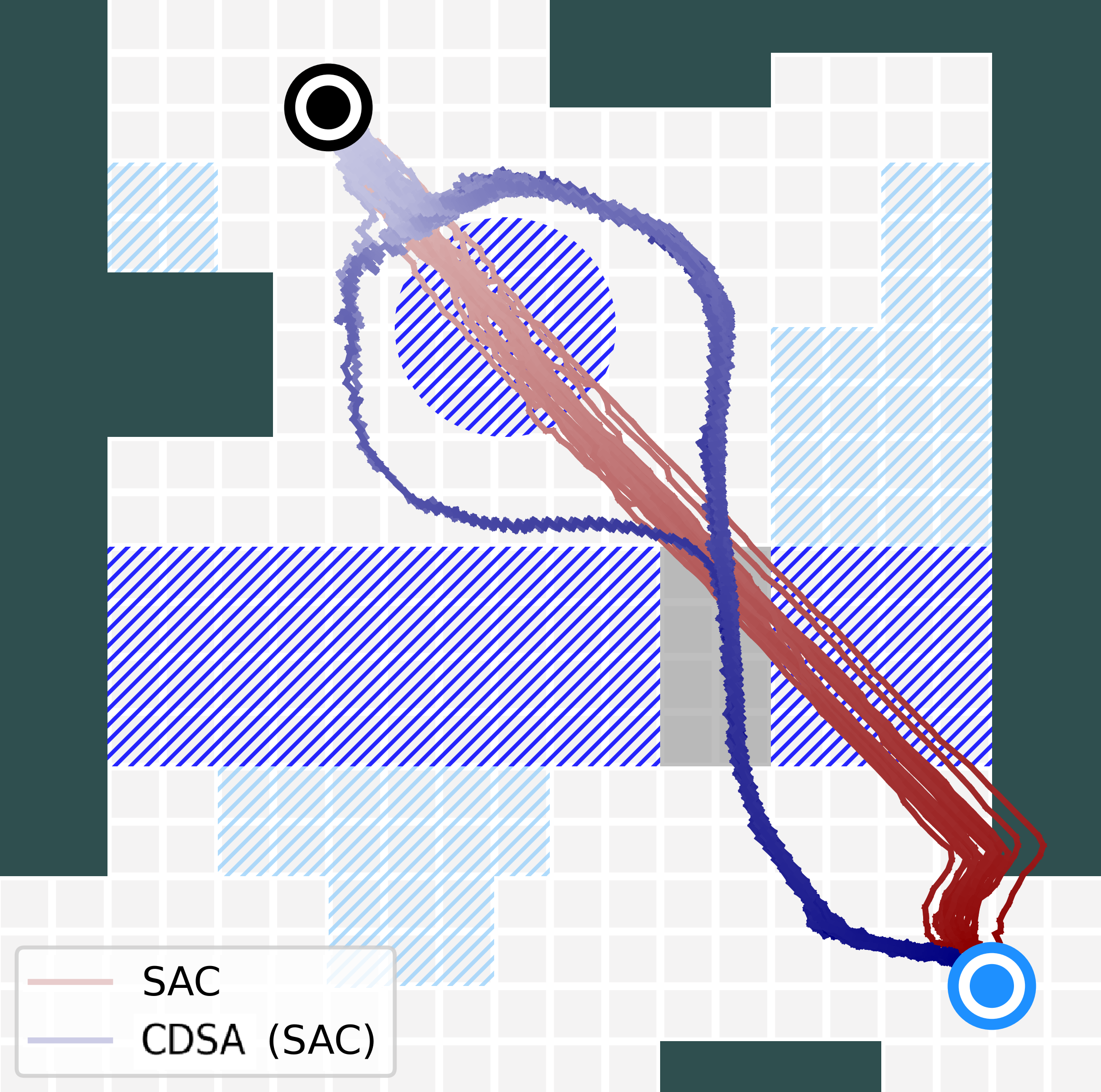}
    \end{minipage}
    }
    \subfigure[]{
    \begin{minipage}[t]{0.32\linewidth}
    \centering
    \includegraphics[height=120pt,width=120pt]{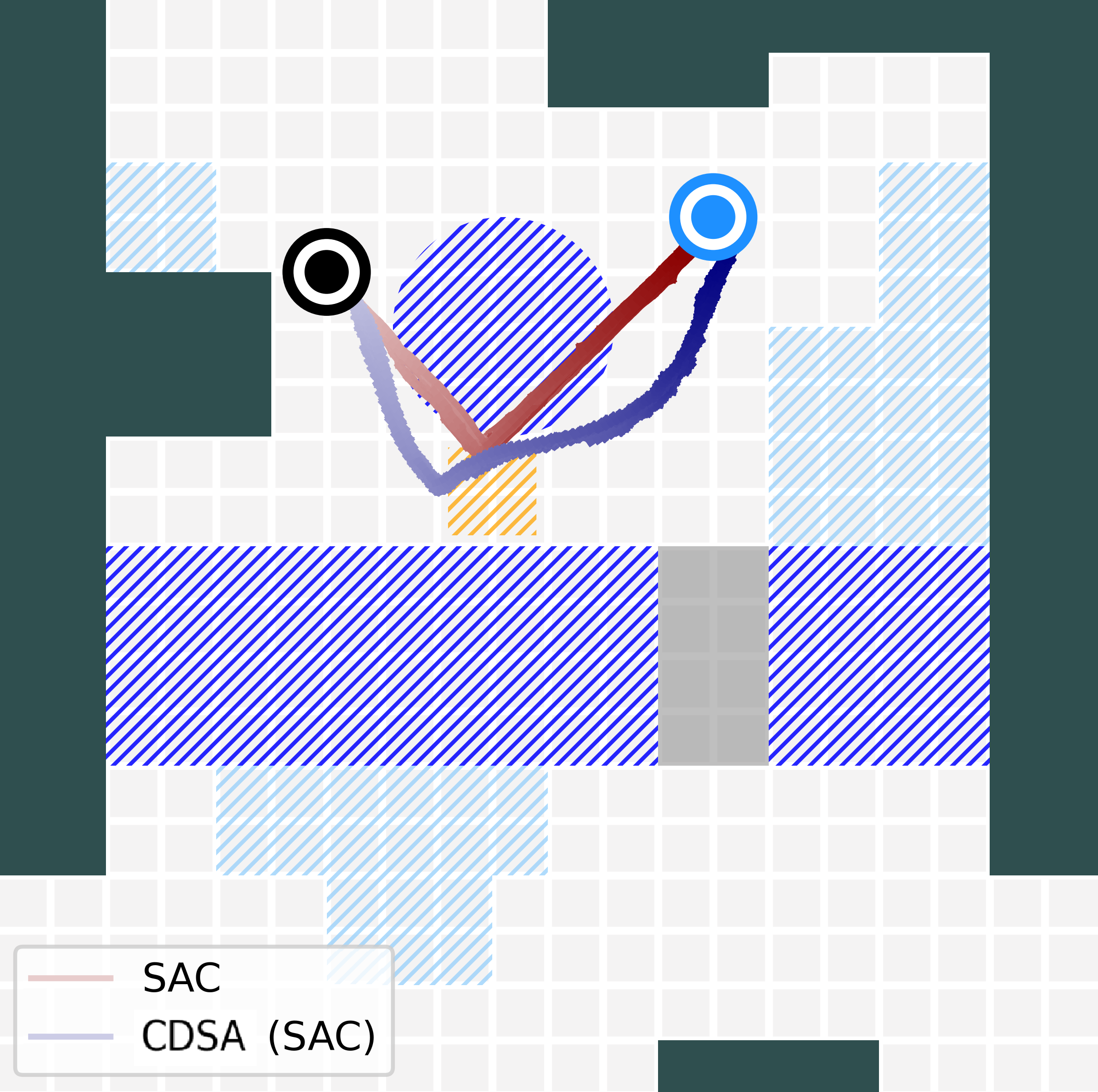}
    \end{minipage}
    }
    \subfigure[]{
    \begin{minipage}[t]{0.32\linewidth}
    \centering
    \includegraphics[height=120pt,width=120pt]{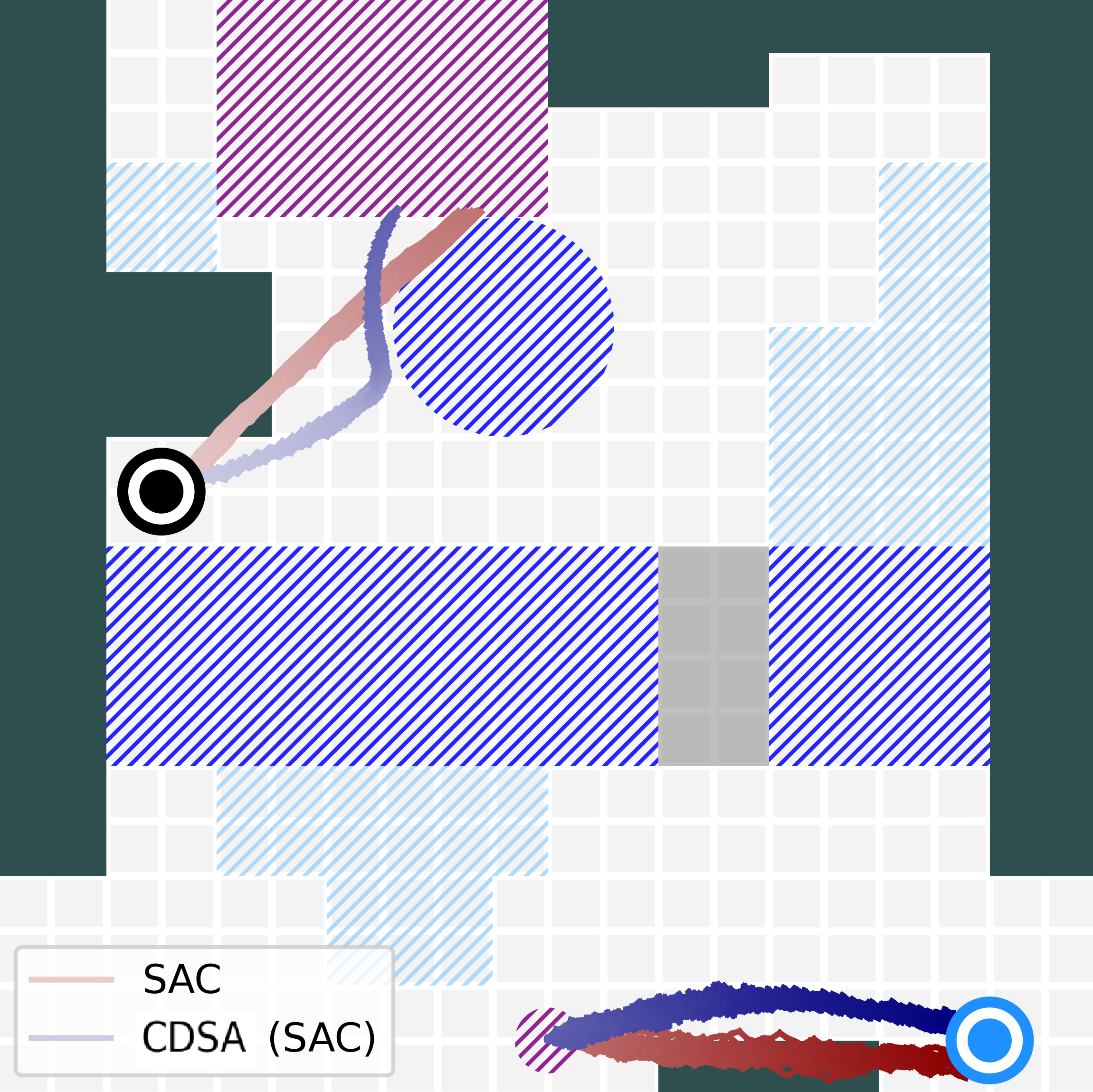}
    \end{minipage}
    }
    \caption{The results of risky transportation experiment. (a) shows the map of the environment. \imgg{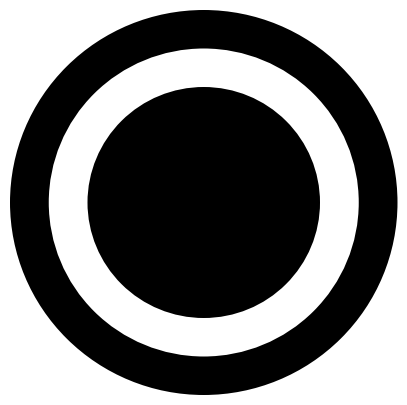} is the start point and \imgg{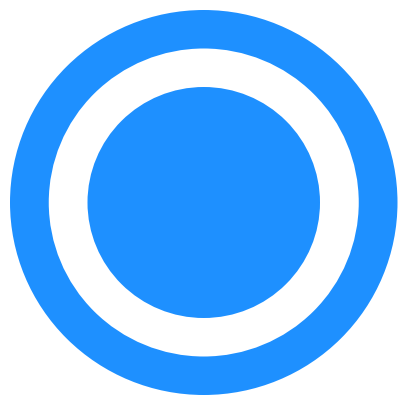} is the target point, the trajectories of the agent are represented by several lines that increase saturation with the number of steps. \imgg{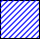} is river, \imgg{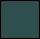} is mountain, and \imgg{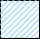} is ice, which are risky regions. (b) shows the states in the offline dataset. The black color represents $\sum_{a} p_{\text{data}}(s,a)=0$ and the white color represents $\sum_{a} p_{\text{data}}(s,a)$ is non-zero in the dataset. (c) is the gradient field of states learned from CDSA. We only show the gradient field of states since the gradient field of actions is hard to present. (d) is the results of SAC and CDSA (SAC). After employing CDSA to maintain the agent within the known region, the agent exhibits a higher degree of risk aversion. (e) is the task that requires the agent to bring goods to the target point, \imgg{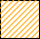} is the region where the goods are placed. (f) shows the results after adding the airport to this environment, \imgg{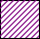} is the airport area. In these two tasks, we use the CDSA models learned from the path finding task without any fine-tuning. The agent avoids all risky regions after using CDSA.}
    \label{FIGURE4}
\end{figure*}

\paragraph{Implementation details.} We run the baseline algorithms IQL and POR with their official codes for 1M gradient steps. For CDSA, we train two score models and the inverse dynamic model on each dataset for 10,000 gradient steps. Due to space limit, we defer the detailed hyperparameter setup as well as the setup of action correction coefficients $K_1, K_2$ on each dataset to table \ref{TABLE3} and table \ref{TABLE4} in the appendix. All algorithms are trained with random seeds 0-4. We report the average performance of each method post-training.

\paragraph{Results.} 
The outcomes for all datasets are summarized in Table \ref{TABLE1}. Significantly, CDSA (POR) and CDSA (IQL) exhibit superior performance across 12 tasks compared to the baseline algorithms, underscoring the efficacy of CDSA in enhancing performance across most datasets. In MuJoCO datasets, CDSA performs exceptionally well in the hopper random dataset (186.1\% for IQL and 155.2\% for POR), with marginal improvements in most datasets (ranging from 0\% to 25.4\%, averaging 6.45\%). We attribute the particular advantage of CDSA in the hopper random dataset to its potentially poor data quality and the relatively simple nature of the hopper controlled by the agents. CDSA's ability to incorporate safety action components aids in stabilizing the robot, thereby yielding higher healthy rewards (reward obtained for maintaining the hopper healthy at each time step). In expert datasets, CDSA shows only limited enhancement, possibly due to the already conservative and high-quality nature of expert datasets, where further conservatism provided by CDSA might not yield substantial improvements. In antmaze datasets, CDSA demonstrates significant improvements across all datasets (ranging from 1.4\% to 40.5\%, averaging 11.4\%), indicating its efficacy in environments with simpler dynamics. We posit that CDSA's success across these datasets stems from the increased likelihood of action-state pairs within the distribution.


\paragraph{Risky-averse Evaluation.} To verify that CDSA can reduce the risk of entering hazardous zones, we visualize the VaR of IQL and CDSA (IQL) as shown in Figure \ref{FIGURE3}. Compared to IQL, CDSA (IQL) improves VaR significantly in almost every percentile. We can also see that VaR increases more when the quantile is smaller since the trajectories of high reward are safe enough and hardly cross risky regions. Overall, the auxiliary action components help the agent take safer action, lowering the probability of risky situations and increasing the cumulative reward in almost every trajectory.

\subsection{Risky transportation}

We use CDSA in the Risky transportation environment. The task of the agent is to find a path from the start point to the target point. We call this task path finding task. There are risky regions such as mountains, rivers, and ice roads, where the agent is at risk of accidents. We give a large negative reward with a small probability in these areas to indicate that the agent has an accident. Before reaching the target point, the agent receives a negative reward proportional to the distance between the agent and the target point.

We use CDSA to learn safe gradient fields of a safe dataset whose trajectory does not contain any risk regions in this environment. We use SAC \cite{haarnoja2018soft} as our baseline algorithms. As depicted in Figure \ref{FIGURE4}, most of the trajectories pass through risky regions to reach the target point quickly. After adding auxiliary safety action components of CDSA in each step, the agent can avoid all risky regions to reach the destination.

To verify that CDSA is effective of generalization, we design two more tasks. The first task is to bring goods to the target point. The agent must reach the point of goods and then go to the target point. In the second task, we add an airport to this environment. Using an airplane, the agent can land near the target point. We only employ the CDSA models trained for the path finding task and do not train any new models for these additional tasks. Figure \ref{FIGURE4} shows that in the first task, the agent finds the goods on the shortest path and travels to the target point despite the presence of risky regions. The agent tends to go to the airport and take a plane to the location near the target point and then goes to the target point in the shortest path in the second task. After adding auxiliary actions generated from CDSA, the trajectories are much safer, and the agent can completely avoid entering risky regions. It is clear that the CDSA model trained in the same environment can be applied to other tasks, which demonstrates the generalization ability of CDSA.


\section{Discussion}
Our work introduces the CDSA algorithm, which learns gradient fields from data and utilizes them to acquire auxiliary actions. These auxiliary actions guide state-action pairs towards high-density regions within the dataset distribution, mitigating exposure to unfamiliar states. Since CDSA focuses solely on learning gradient fields from data, independent of RL baseline algorithms, it seamlessly integrates with various algorithms such as CQL, IQL, and POR. Our experiments in offline settings demonstrate that our method effectively navigates away from hazardous areas and makes decisions within familiar scenarios within the dataset distribution. Combining baseline algorithms with CDSA leads to improved performance on D4RL datasets across various qualities. Notably, CDSA (IQL) and CDSA (POR) exhibit superior performance in 12 tasks. Furthermore, employing CDSA significantly enhances the Value at Risk (VaR) of baseline algorithms, underscoring our method's risk-averse capability. In the Risky Transportation environment, we visualize and validate the generalizability of CDSA across different tasks within the same environment.

While our method shows promising results, it is essential to acknowledge its limitations. Specifically, CDSA's effectiveness is confined to scenarios with continuous action spaces, limiting its applicability in discrete action spaces. Additionally, accurately determining hyperparameters $K_1$ and $K_2$ poses a challenge, requiring careful balancing to achieve optimal performance. However, setting these hyperparameters to the same value often yields satisfactory results, reducing the need for extensive tuning efforts. An area for future exploration involves developing an automated mechanism for adjusting $K_1$ and $K_2$.

\begin{ack}
By using the \texttt{ack} environment to insert your (optional) 
acknowledgements, you can ensure that the text is suppressed whenever 
you use the \texttt{doubleblind} option. In the final version, 
acknowledgements may be included on the extra page intended for references.
\end{ack}



\bibliography{mybibfile}

\begin{thebibliography}{52}
\providecommand{\natexlab}[1]{#1}
\providecommand{\url}[1]{\texttt{#1}}
\expandafter\ifx\csname urlstyle\endcsname\relax
  \providecommand{\doi}[1]{doi: #1}\else
  \providecommand{\doi}{doi: \begingroup \urlstyle{rm}\Url}\fi

\bibitem[Abbeel and Ng(2004)]{abbeel2004apprenticeship}
P.~Abbeel and A.~Y. Ng.
\newblock Apprenticeship learning via inverse reinforcement learning.
\newblock In \emph{Proceedings of the twenty-first international conference on
  Machine learning}, page~1, 2004.

\bibitem[An et~al.(2021)An, Moon, Kim, and Song]{an2021uncertainty}
G.~An, S.~Moon, J.-H. Kim, and H.~O. Song.
\newblock Uncertainty-based offline reinforcement learning with diversified
  q-ensemble.
\newblock \emph{Advances in neural information processing systems},
  34:\penalty0 7436--7447, 2021.

\bibitem[Bai et~al.(2022)Bai, Wang, Yang, Deng, Garg, Liu, and
  Wang]{bai2022pessimistic}
C.~Bai, L.~Wang, Z.~Yang, Z.~Deng, A.~Garg, P.~Liu, and Z.~Wang.
\newblock Pessimistic bootstrapping for uncertainty-driven offline
  reinforcement learning.
\newblock \emph{arXiv preprint arXiv:2202.11566}, 2022.

\bibitem[Brandfonbrener et~al.(2021)Brandfonbrener, Whitney, Ranganath, and
  Bruna]{brandfonbrener2021offline}
D.~Brandfonbrener, W.~Whitney, R.~Ranganath, and J.~Bruna.
\newblock Offline rl without off-policy evaluation.
\newblock \emph{Advances in Neural Information Processing Systems},
  34:\penalty0 4933--4946, 2021.

\bibitem[Chen et~al.(2021)Chen, Lu, Rajeswaran, Lee, Grover, Laskin, Abbeel,
  Srinivas, and Mordatch]{chen2021decision}
L.~Chen, K.~Lu, A.~Rajeswaran, K.~Lee, A.~Grover, M.~Laskin, P.~Abbeel,
  A.~Srinivas, and I.~Mordatch.
\newblock Decision transformer: Reinforcement learning via sequence modeling.
\newblock \emph{Advances in neural information processing systems},
  34:\penalty0 15084--15097, 2021.

\bibitem[Diehl et~al.(2021)Diehl, Sievernich, Kr{\"u}ger, Hoffmann, and
  Bertran]{diehl2021umbrella}
C.~Diehl, T.~Sievernich, M.~Kr{\"u}ger, F.~Hoffmann, and T.~Bertran.
\newblock Umbrella: Uncertainty-aware model-based offline reinforcement
  learning leveraging planning.
\newblock \emph{arXiv preprint arXiv:2111.11097}, 2021.

\bibitem[Fu et~al.(2020)Fu, Kumar, Nachum, Tucker, and Levine]{fu2020d4rl}
J.~Fu, A.~Kumar, O.~Nachum, G.~Tucker, and S.~Levine.
\newblock D4rl: Datasets for deep data-driven reinforcement learning.
\newblock \emph{arXiv preprint arXiv:2004.07219}, 2020.

\bibitem[Fujimoto and Gu(2021)]{fujimoto2021minimalist}
S.~Fujimoto and S.~S. Gu.
\newblock A minimalist approach to offline reinforcement learning.
\newblock \emph{Advances in neural information processing systems},
  34:\penalty0 20132--20145, 2021.

\bibitem[Fujimoto et~al.(2019)Fujimoto, Meger, and Precup]{fujimoto2019off}
S.~Fujimoto, D.~Meger, and D.~Precup.
\newblock Off-policy deep reinforcement learning without exploration.
\newblock In \emph{International conference on machine learning}, pages
  2052--2062. PMLR, 2019.

\bibitem[Gelada and Bellemare(2019)]{gelada2019off}
C.~Gelada and M.~G. Bellemare.
\newblock Off-policy deep reinforcement learning by bootstrapping the covariate
  shift.
\newblock In \emph{Proceedings of the AAAI Conference on Artificial
  Intelligence}, volume~33, pages 3647--3655, 2019.

\bibitem[Haarnoja et~al.(2018)Haarnoja, Zhou, Abbeel, and
  Levine]{haarnoja2018soft}
T.~Haarnoja, A.~Zhou, P.~Abbeel, and S.~Levine.
\newblock Soft actor-critic: Off-policy maximum entropy deep reinforcement
  learning with a stochastic actor.
\newblock In \emph{International conference on machine learning}, pages
  1861--1870. PMLR, 2018.

\bibitem[Ho and Ermon(2016)]{ho2016generative}
J.~Ho and S.~Ermon.
\newblock Generative adversarial imitation learning.
\newblock \emph{Advances in neural information processing systems}, 29, 2016.

\bibitem[Ho et~al.(2020)Ho, Jain, and Abbeel]{ho2020denoising}
J.~Ho, A.~Jain, and P.~Abbeel.
\newblock Denoising diffusion probabilistic models.
\newblock \emph{Advances in Neural Information Processing Systems},
  33:\penalty0 6840--6851, 2020.

\bibitem[Hyv{\"a}rinen and Dayan(2005)]{hyvarinen2005estimation}
A.~Hyv{\"a}rinen and P.~Dayan.
\newblock Estimation of non-normalized statistical models by score matching.
\newblock \emph{Journal of Machine Learning Research}, 6\penalty0 (4), 2005.

\bibitem[Kang et~al.(2022)Kang, Gradu, Choi, Janner, Tomlin, and
  Levine]{kang2022lyapunov}
K.~Kang, P.~Gradu, J.~Choi, M.~Janner, C.~Tomlin, and S.~Levine.
\newblock Lyapunov density models: Constraining distribution shift in
  learning-based control, 2022.

\bibitem[Kong et~al.(2020)Kong, Ping, Huang, Zhao, and
  Catanzaro]{kong2020diffwave}
Z.~Kong, W.~Ping, J.~Huang, K.~Zhao, and B.~Catanzaro.
\newblock Diffwave: A versatile diffusion model for audio synthesis.
\newblock \emph{arXiv preprint arXiv:2009.09761}, 2020.

\bibitem[Kostrikov et~al.(2021)Kostrikov, Nair, and
  Levine]{kostrikov2021offline}
I.~Kostrikov, A.~Nair, and S.~Levine.
\newblock Offline reinforcement learning with implicit q-learning.
\newblock \emph{arXiv preprint arXiv:2110.06169}, 2021.

\bibitem[Kumar et~al.(2019)Kumar, Fu, Soh, Tucker, and
  Levine]{kumar2019stabilizing}
A.~Kumar, J.~Fu, M.~Soh, G.~Tucker, and S.~Levine.
\newblock Stabilizing off-policy q-learning via bootstrapping error reduction.
\newblock \emph{Advances in Neural Information Processing Systems}, 32, 2019.

\bibitem[Kumar et~al.(2020)Kumar, Zhou, Tucker, and
  Levine]{kumar2020conservative}
A.~Kumar, A.~Zhou, G.~Tucker, and S.~Levine.
\newblock Conservative q-learning for offline reinforcement learning.
\newblock \emph{Advances in Neural Information Processing Systems},
  33:\penalty0 1179--1191, 2020.

\bibitem[Lange et~al.(2012)Lange, Gabel, and Riedmiller]{lange2012batch}
S.~Lange, T.~Gabel, and M.~Riedmiller.
\newblock Batch reinforcement learning.
\newblock In \emph{Reinforcement learning}, pages 45--73. Springer, 2012.

\bibitem[Levine et~al.(2020)Levine, Kumar, Tucker, and Fu]{levine2020offline}
S.~Levine, A.~Kumar, G.~Tucker, and J.~Fu.
\newblock Offline reinforcement learning: Tutorial, review, and perspectives on
  open problems.
\newblock \emph{arXiv preprint arXiv:2005.01643}, 2020.

\bibitem[Liu et~al.(2016)Liu, Lee, and Jordan]{liu2016kernelized}
Q.~Liu, J.~Lee, and M.~Jordan.
\newblock A kernelized stein discrepancy for goodness-of-fit tests.
\newblock In \emph{International conference on machine learning}, pages
  276--284. PMLR, 2016.

\bibitem[Liu et~al.(2019)Liu, Swaminathan, Agarwal, and Brunskill]{liu2019off}
Y.~Liu, A.~Swaminathan, A.~Agarwal, and E.~Brunskill.
\newblock Off-policy policy gradient with state distribution correction.
\newblock \emph{arXiv preprint arXiv:1904.08473}, 2019.

\bibitem[Lyu et~al.(2022{\natexlab{a}})Lyu, Li, and Lu]{lyu2022doublecheck}
J.~Lyu, X.~Li, and Z.~Lu.
\newblock Double check your state before trusting it: Confidence-aware
  bidirectional offline model-based imagination.
\newblock In \emph{Thirty-sixth Conference on Neural Information Processing
  Systems}, 2022{\natexlab{a}}.

\bibitem[Lyu et~al.(2022{\natexlab{b}})Lyu, Ma, Li, and Lu]{lyu2022mildly}
J.~Lyu, X.~Ma, X.~Li, and Z.~Lu.
\newblock Mildly conservative q-learning for offline reinforcement learning.
\newblock In \emph{Thirty-sixth Conference on Neural Information Processing
  Systems}, 2022{\natexlab{b}}.

\bibitem[Lyu et~al.(2023)Lyu, Gong, Wan, Lu, and Li]{lyu2023state}
J.~Lyu, A.~Gong, L.~Wan, Z.~Lu, and X.~Li.
\newblock State advantage weighting for offline {RL}.
\newblock In \emph{International Conference on Learning Representation tiny
  paper}, 2023.
\newblock URL \url{https://openreview.net/forum?id=PjypHLTo29v}.

\bibitem[Ma et~al.(2021)Ma, Jayaraman, and Bastani]{ma2021conservative}
Y.~Ma, D.~Jayaraman, and O.~Bastani.
\newblock Conservative offline distributional reinforcement learning.
\newblock \emph{Advances in Neural Information Processing Systems},
  34:\penalty0 19235--19247, 2021.

\bibitem[McAllister et~al.(2019)McAllister, Kahn, Clune, and
  Levine]{mcallister2019robustness}
R.~McAllister, G.~Kahn, J.~Clune, and S.~Levine.
\newblock Robustness to out-of-distribution inputs via task-aware generative
  uncertainty.
\newblock In \emph{2019 International Conference on Robotics and Automation
  (ICRA)}, pages 2083--2089. IEEE, 2019.

\bibitem[Mnih et~al.(2013)Mnih, Kavukcuoglu, Silver, Graves, Antonoglou,
  Wierstra, and Riedmiller]{mnih2013playing}
V.~Mnih, K.~Kavukcuoglu, D.~Silver, A.~Graves, I.~Antonoglou, D.~Wierstra, and
  M.~Riedmiller.
\newblock Playing atari with deep reinforcement learning.
\newblock \emph{arXiv preprint arXiv:1312.5602}, 2013.

\bibitem[Namkoong and Duchi(2017)]{namkoong2017variance}
H.~Namkoong and J.~C. Duchi.
\newblock Variance-based regularization with convex objectives.
\newblock \emph{Advances in neural information processing systems}, 30, 2017.

\bibitem[Niu et~al.(2020)Niu, Song, Song, Zhao, Grover, and
  Ermon]{niu2020permutation}
C.~Niu, Y.~Song, J.~Song, S.~Zhao, A.~Grover, and S.~Ermon.
\newblock Permutation invariant graph generation via score-based generative
  modeling.
\newblock In \emph{International Conference on Artificial Intelligence and
  Statistics}, pages 4474--4484. PMLR, 2020.

\bibitem[Richter and Roy(2017)]{richter2017safe}
C.~Richter and N.~Roy.
\newblock Safe visual navigation via deep learning and novelty detection.
\newblock 2017.

\bibitem[Rockafellar and Uryasev(2002)]{rockafellar2002conditional}
R.~T. Rockafellar and S.~Uryasev.
\newblock Conditional value-at-risk for general loss distributions.
\newblock \emph{Journal of banking \& finance}, 26\penalty0 (7):\penalty0
  1443--1471, 2002.

\bibitem[Schulman et~al.(2015)Schulman, Levine, Abbeel, Jordan, and
  Moritz]{schulman2015trust}
J.~Schulman, S.~Levine, P.~Abbeel, M.~Jordan, and P.~Moritz.
\newblock Trust region policy optimization.
\newblock In \emph{International conference on machine learning}, pages
  1889--1897. PMLR, 2015.

\bibitem[Silver et~al.(2016)Silver, Huang, Maddison, Guez, Sifre, Van
  Den~Driessche, Schrittwieser, Antonoglou, Panneershelvam, Lanctot,
  et~al.]{silver2016mastering}
D.~Silver, A.~Huang, C.~J. Maddison, A.~Guez, L.~Sifre, G.~Van Den~Driessche,
  J.~Schrittwieser, I.~Antonoglou, V.~Panneershelvam, M.~Lanctot, et~al.
\newblock Mastering the game of go with deep neural networks and tree search.
\newblock \emph{nature}, 529\penalty0 (7587):\penalty0 484--489, 2016.

\bibitem[Song and Ermon(2019)]{song2019generative}
Y.~Song and S.~Ermon.
\newblock Generative modeling by estimating gradients of the data distribution.
\newblock \emph{Advances in Neural Information Processing Systems}, 32, 2019.

\bibitem[Song et~al.(2020)Song, Sohl-Dickstein, Kingma, Kumar, Ermon, and
  Poole]{song2020score}
Y.~Song, J.~Sohl-Dickstein, D.~P. Kingma, A.~Kumar, S.~Ermon, and B.~Poole.
\newblock Score-based generative modeling through stochastic differential
  equations.
\newblock \emph{arXiv preprint arXiv:2011.13456}, 2020.

\bibitem[Sutton et~al.(1998)Sutton, Barto, et~al.]{sutton1998introduction}
R.~S. Sutton, A.~G. Barto, et~al.
\newblock Introduction to reinforcement learning.
\newblock 1998.

\bibitem[Sutton et~al.(2016)Sutton, Mahmood, and White]{sutton2016emphatic}
R.~S. Sutton, A.~R. Mahmood, and M.~White.
\newblock An emphatic approach to the problem of off-policy temporal-difference
  learning.
\newblock \emph{The Journal of Machine Learning Research}, 17\penalty0
  (1):\penalty0 2603--2631, 2016.

\bibitem[Tversky and Kahneman(1992)]{tversky1992advances}
A.~Tversky and D.~Kahneman.
\newblock Advances in prospect theory: Cumulative representation of
  uncertainty.
\newblock \emph{Journal of Risk and uncertainty}, 5\penalty0 (4):\penalty0
  297--323, 1992.

\bibitem[Urp{\'\i} et~al.(2021)Urp{\'\i}, Curi, and Krause]{urpi2021risk}
N.~A. Urp{\'\i}, S.~Curi, and A.~Krause.
\newblock Risk-averse offline reinforcement learning.
\newblock \emph{arXiv preprint arXiv:2102.05371}, 2021.

\bibitem[Vahdat et~al.(2021)Vahdat, Kreis, and Kautz]{vahdat2021score}
A.~Vahdat, K.~Kreis, and J.~Kautz.
\newblock Score-based generative modeling in latent space.
\newblock \emph{Advances in Neural Information Processing Systems},
  34:\penalty0 11287--11302, 2021.

\bibitem[Vincent(2011)]{vincent2011connection}
P.~Vincent.
\newblock A connection between score matching and denoising autoencoders.
\newblock \emph{Neural computation}, 23\penalty0 (7):\penalty0 1661--1674,
  2011.

\bibitem[Wang(1996)]{wang1996premium}
S.~Wang.
\newblock Premium calculation by transforming the layer premium density.
\newblock \emph{ASTIN Bulletin: The Journal of the IAA}, 26\penalty0
  (1):\penalty0 71--92, 1996.

\bibitem[Wang et~al.(2022)Wang, Hunt, and Zhou]{wang2022diffusion}
Z.~Wang, J.~J. Hunt, and M.~Zhou.
\newblock Diffusion policies as an expressive policy class for offline
  reinforcement learning.
\newblock \emph{arXiv preprint arXiv:2208.06193}, 2022.

\bibitem[Welling and Teh(2011)]{welling2011bayesian}
M.~Welling and Y.~W. Teh.
\newblock Bayesian learning via stochastic gradient langevin dynamics.
\newblock In \emph{Proceedings of the 28th international conference on machine
  learning (ICML-11)}, pages 681--688, 2011.

\bibitem[Wu et~al.(2022)Wu, Zhong, Xia, and Dong]{wu2022targf}
M.~Wu, F.~Zhong, Y.~Xia, and H.~Dong.
\newblock Targf: Learning target gradient field for object rearrangement.
\newblock \emph{arXiv preprint arXiv:2209.00853}, 2022.

\bibitem[Wu et~al.(2021)Wu, Zhai, Srivastava, Susskind, Zhang, Salakhutdinov,
  and Goh]{wu2021uncertainty}
Y.~Wu, S.~Zhai, N.~Srivastava, J.~Susskind, J.~Zhang, R.~Salakhutdinov, and
  H.~Goh.
\newblock Uncertainty weighted actor-critic for offline reinforcement learning.
\newblock \emph{arXiv preprint arXiv:2105.08140}, 2021.

\bibitem[Xu et~al.(2022)Xu, Jiang, Li, and Zhan]{xu2022policy}
H.~Xu, L.~Jiang, J.~Li, and X.~Zhan.
\newblock A policy-guided imitation approach for offline reinforcement
  learning.
\newblock \emph{arXiv preprint arXiv:2210.08323}, 2022.

\bibitem[Yang et~al.(2024)Yang, Tao, Lyu, and Li]{yang2024exploration}
K.~Yang, J.~Tao, J.~Lyu, and X.~Li.
\newblock Exploration and anti-exploration with distributional random network
  distillation.
\newblock \emph{arXiv preprint arXiv:2401.09750}, 2024.

\bibitem[Yu et~al.(2020)Yu, Thomas, Yu, Ermon, Zou, Levine, Finn, and
  Ma]{yu2020mopo}
T.~Yu, G.~Thomas, L.~Yu, S.~Ermon, J.~Y. Zou, S.~Levine, C.~Finn, and T.~Ma.
\newblock Mopo: Model-based offline policy optimization.
\newblock \emph{Advances in Neural Information Processing Systems},
  33:\penalty0 14129--14142, 2020.

\bibitem[Zhang et~al.(2023)Zhang, Lyu, Ma, Yan, Yang, Wan, and
  Li]{Zhang2023UncertaintydrivenTT}
J.~Zhang, J.~Lyu, X.~Ma, J.~Yan, J.~Yang, L.~Wan, and X.~Li.
\newblock Uncertainty-driven trajectory truncation for model-based offline
  reinforcement learning.
\newblock \emph{ArXiv}, abs/2304.04660, 2023.

\end{thebibliography}

\appendix

\section{Proof}
\subsection{Proof of equivalent loss form}
\label{section:A.1}
The loss of network $g_{\theta}(\tilde{x})$ is
$$
\mathcal{L}_\theta=\frac{1}{2} \mathbb{E}_{q_{\sigma}(\tilde{\mathbf{x}} \mid \mathbf{x}) p_{\text {data }}(\mathbf{x})}\left[\left\|g_{\theta}(\tilde{\mathbf{x}})-\nabla_{\tilde{a}} \log q_{\sigma}(\tilde{\mathbf{x}} \mid \mathbf{x})\right\|_{2}^{2}\right].
$$
When the noise distribution $q_\sigma(\tilde{\mathbf{x}})$ obey $N(\mathbf{x},\sigma I)$, the partial derivative of $\log q_\sigma(\tilde{\mathbf{x}})$ with respect to $a$ is

\begin{align*}
\nabla_{\tilde{a}} \log q_{\sigma}(\tilde{\mathbf{x}} \mid \mathbf{x}) 
& = \nabla_{\tilde{a}} \log \left(\frac{1}{\sqrt{2 \pi} \sigma} e^{-\frac{(\tilde{\mathbf{x}}-{x})^{2}}{2 \sigma^{2}}}\right) \\ & = \nabla_{\tilde{a}}\left(C-\frac{(\tilde{\mathbf{x}}-\mathbf{x})^{2}}{2 \sigma^{2}}\right) \\ & = -\frac{\partial \tilde{\mathbf{x}}}{\partial \tilde{a}} \frac{(\tilde{\mathbf{x}}-\mathbf{x})}{\sigma^{2}} \\ & = -(0,1) \cdot\left(\frac{(\tilde{\mathbf{s}}-\mathbf{s})}{\sigma^{2}}, \frac{(\tilde{\mathbf{a}}-\mathbf{a})}{\sigma^{2}}\right) \\ & = -\frac{(\tilde{\mathbf{a}}-\mathbf{a})}{\sigma^{2}}. \\
\end{align*}

Using reparameterization trick, $\mathbf{\tilde{a}}$ can be expressed as $\mathbf{\tilde{a}}=\mathbf{a}+\sigma z$ where $z$ follows the standard normal distribution.
$\mathbf{\tilde{x}}$ can be also represent as $\mathbf{\tilde{x}}=\mathbf{x}+(0,\sigma z)$. The loss formula can be converted into
\begin{align*}
\mathcal{L}_\theta &=\frac{1}{2} \mathbb{E}_{q_{\sigma}(\tilde{\mathbf{x}} \mid \mathbf{x}) p_{\text {data }}(\mathbf{x})}\left[\left\|g_{\theta}(\tilde{\mathbf{x}})-\nabla_{\tilde{a}} \log q_{\sigma}(\tilde{\mathbf{x}} \mid \mathbf{x})\right\|_{2}^{2}\right]\\
&=\frac{1}{2} \mathbb{E}_{q_{\sigma}(\tilde{\mathbf{x}} \mid \mathbf{x}) p_{\text {data }}(\mathbf{x})}\left[\left\|g_{\theta}(\tilde{\mathbf{x}})+\frac{(\tilde{\mathbf{a}}-\mathbf{a})}{\sigma^2}\right\|_{2}^{2}\right]\\
&=\frac{1}{2} \mathbb{E}_{q_{\sigma}(\tilde{\mathbf{x}} \mid \mathbf{x}) p_{\text {data }}(\mathbf{x})}\left[\left\|g_{\theta}({\mathbf{x}+(0,\sigma z)})+\frac{z}{\sigma}\right\|_{2}^{2}\right]\\
&=\frac{1}{2} \mathbb{E}_{q_{\sigma}(\tilde{\mathbf{x}} \mid \mathbf{x}) p_{\text {data }}(\mathbf{x})}\left[\left\|g_{\theta}(\mathbf{x}+\sigma\mathbf{z})+\frac{z}{\sigma}\right\|_{2}^{2}\right].
\end{align*}
where $\mathbf{z}=(0,z)$ and $z \sim N(0,I)$.

\section{Experiment details}
Our experiments were performed by using the following hardware and software:
\begin{itemize}
    \item GPUs: NVIDIA GeForce RTX 3090
    \item Python 3.10.8
    \item Numpy 1.23.4
    \item Pytorch 1.13.0
    \item Mujoco-py 2.1.2.14
    \item Mujoco 2.3.1
    \item D4RL 1.1
\end{itemize}

\subsection{D4RL experiments}
\begin{figure*}
    \centering
    \includegraphics[height=640pt,width=440pt]{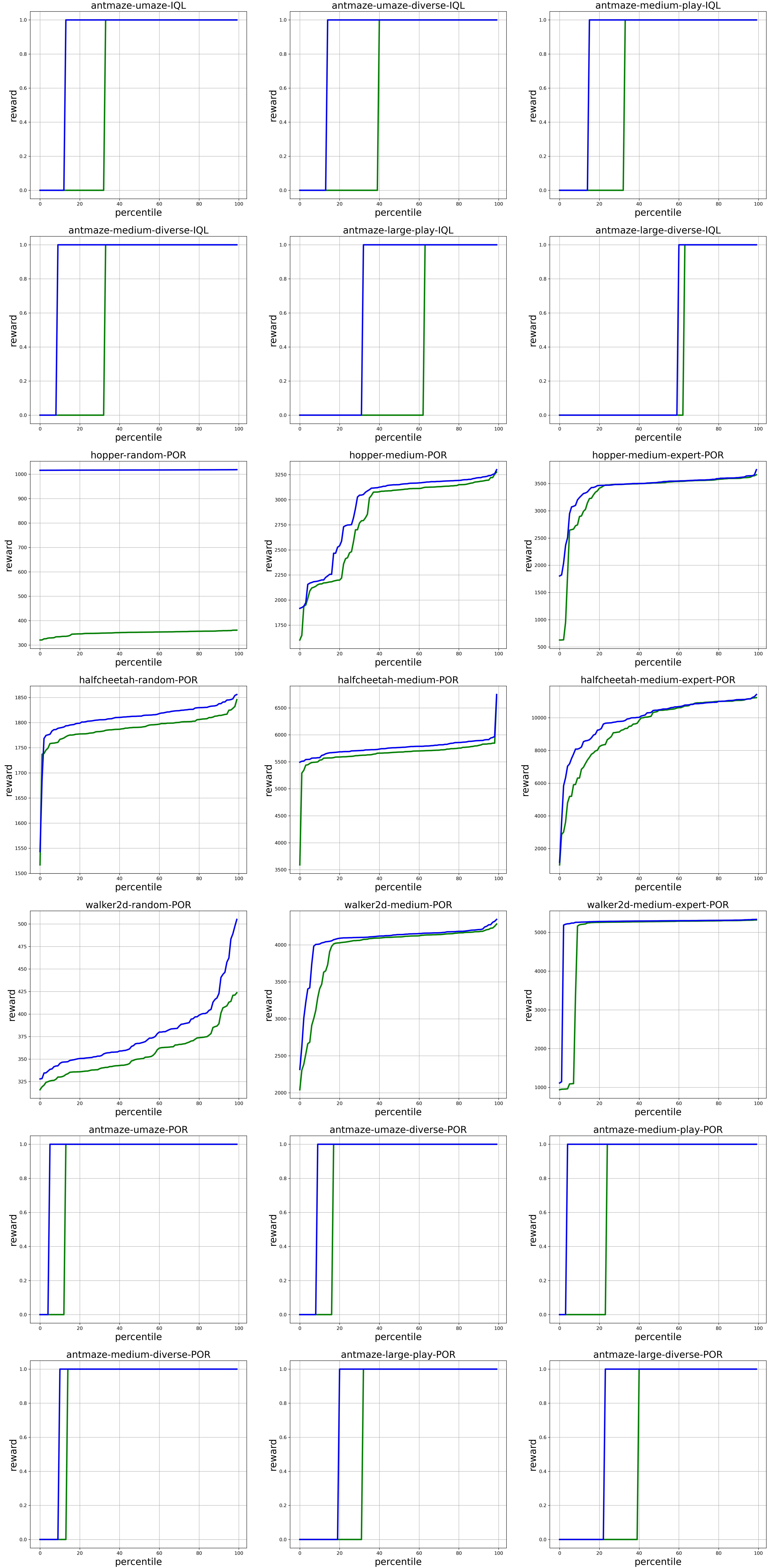}
    \caption{Results of CDSA (POR) and POR.}
    \label{FIGURE5}
\end{figure*}

The VaR of POR and CDSA(POR) are shown in Figure \ref{FIGURE5}. The blue lines represent the result of the baseline algorithm with CDSA and the green lines show the result of the baseline algorithm. We can see that CDSA can improve the VaR in almost every dataset, especially when the percentile is small, verifying the risk-averse effect of CDSA.

\label{section:B.1}

\begin{table}[h]
    \centering
\caption{Hyperparameters of each dataset}
\label{TABLE4}
        \begin{tabular}{l| l| l| l| l}
        \toprule
        &\multicolumn{2}{c|}{\textbf{IQL}} &\multicolumn{2}{c}{\textbf{POR}} \\
         \cline{2-5} & K1 & K2 & K1 & K2\\
        \hline
          hopper-r & 0.1 & 1 &0.1&0.5\\ 
           hopper-m & 0.001 & 0.001 &0.001&0.001\\
            hopper-m-e & 0.001&0.001 &0.001&0.001\\
            \hline
            halfcheetah-r & 0.001&0.005 &0.003&0.03\\
            halfcheetah-m & 0.005 & 0.05 &0.01&0.01\\
            halfcheetah-m-e & 0.01 & 0.01 &0.003&0.03\\
            \hline
            walker2d-r & 0.03 & 0.3 &0.00&0.01\\
            walker2d-m & 0.03 & 0.003 &0.005&0.01\\
            walker2d-m-e & 0.1 & 0.01 &0.00&0.01\\
            \hline
            antmaze-u & 0.5 & 0.1 &0.1&0.1\\
            antmaze-u-d & 0.1 & 0.5 &0.05&0.05\\
            antmaze-m-p & 0.1 & 0.3 &0.1&0.1\\
            antmaze-m-d & 0.3 & 0.3 &0.05&0.05\\
            antmaze-l-p & 0.3 & 0.3 &0.1&0.1\\
            antmaze-l-d & 0.1 & 0.1 &0.03&0.03\\    
         \bottomrule 
        \end{tabular}
\end{table}

\begin{table}[h]
\caption{Hyperparameters of CDSA}
    \centering
\label{TABLE3}
\resizebox{\linewidth}{!}{
    \begin{tabular}{l l l}
        \toprule
        & \textbf{Name} & \textbf{Value} \\
        \midrule
          \multirow{9}{*}{Architecture}& Hidden layers of $g_{\theta}$ & 3\\ 
          &   Hidden dim of $g_{\theta}$ & 32,128,32\\
         &   Activation function of $g_{\theta}$ & LeakyReLU(0.1)\\
         &   Hidden layers of $h_{\varphi}$ & 3\\ 
          &   Hidden dims of $h_{\varphi}$ & 32,128,32\\
         &   Activation function of $h_{\varphi}$& LeakyReLU(0.1)\\
       &   Hidden layers of $I_{\phi}$ & 3\\ 
          &   Hidden dims of $I_{\phi}$ & 128\\
         &   Activation function of $I_{\phi}$ & LeakyReLU(0.2)\\
         \hline \multirow{6}{*}{Hyperparameters} & Optimizer & Adam\\
         &   Learning rate of $g_{\theta}$& 3e-4\\
         &   Learning rate of $h_{\varphi}$& 3e-4\\
         &   Learning rate of $I_{\phi}$& 1e-3\\
         & Batch size & 256\\
         & Iteration number & 10000\\
         
        \bottomrule 
        \end{tabular}
        }
\end{table}

\section{Ablation Study}
\label{section:C}
\begin{figure*}[h]
    \centering
    \subfigure{
    \begin{minipage}[t]{0.3\linewidth}
    \centering
    \includegraphics[height=80pt,width=155pt]{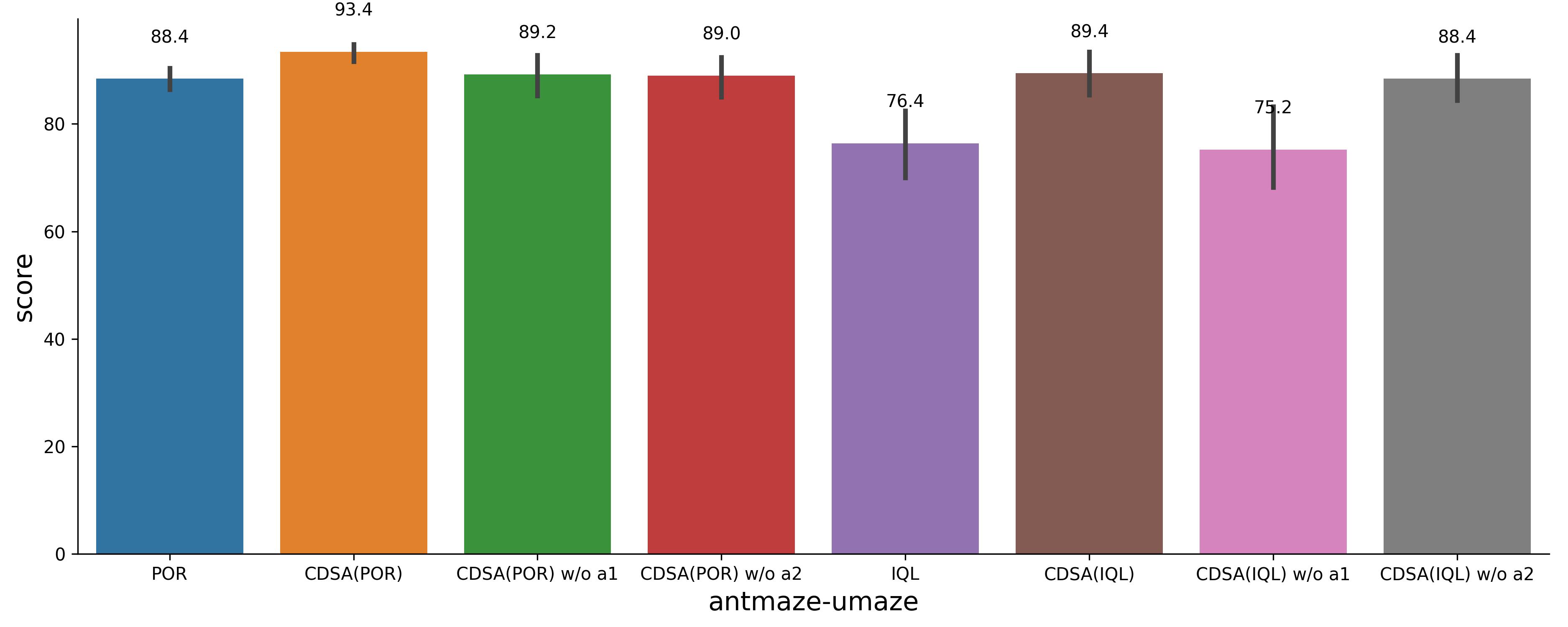}
    \end{minipage}
    }
    \subfigure{
    \begin{minipage}[t]{0.3\linewidth}
    \centering
    \includegraphics[height=80pt,width=155pt]{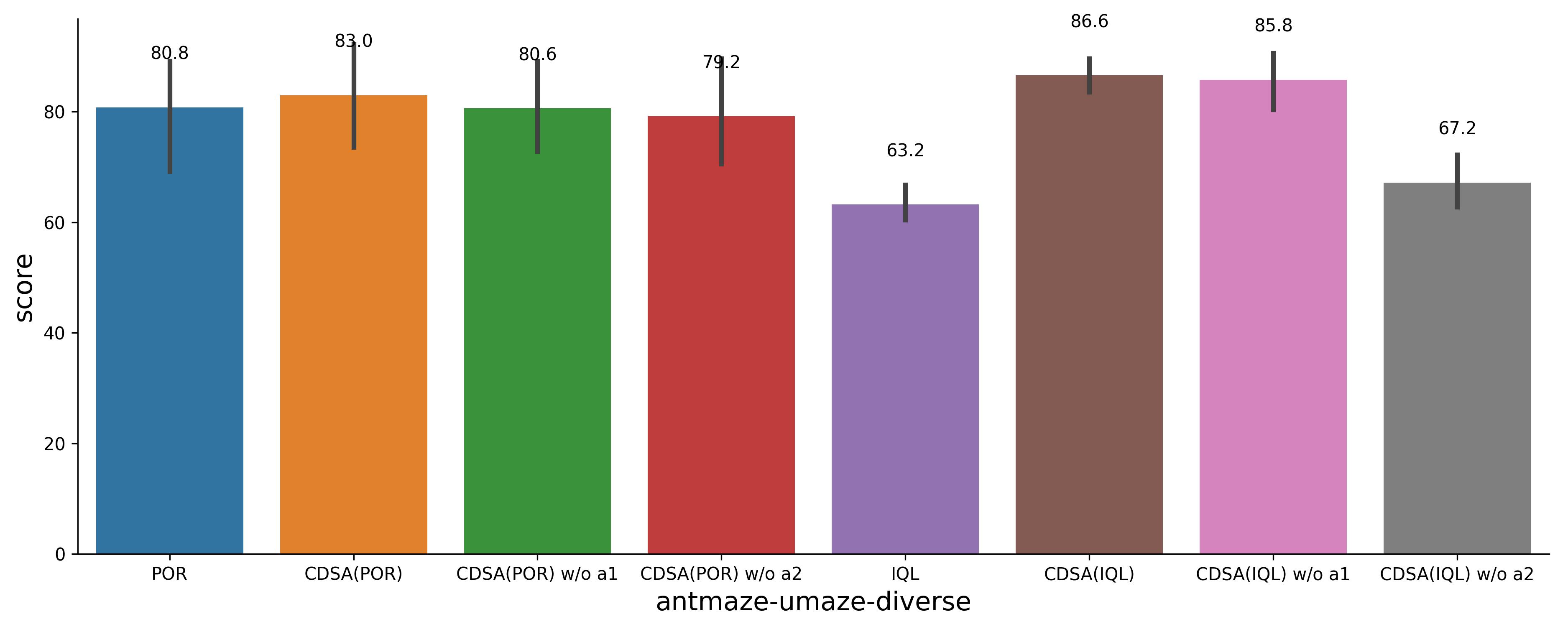}
    \end{minipage}
    }
    \subfigure{
    \begin{minipage}[t]{0.3\linewidth}
    \centering
    \includegraphics[height=80pt,width=155pt]{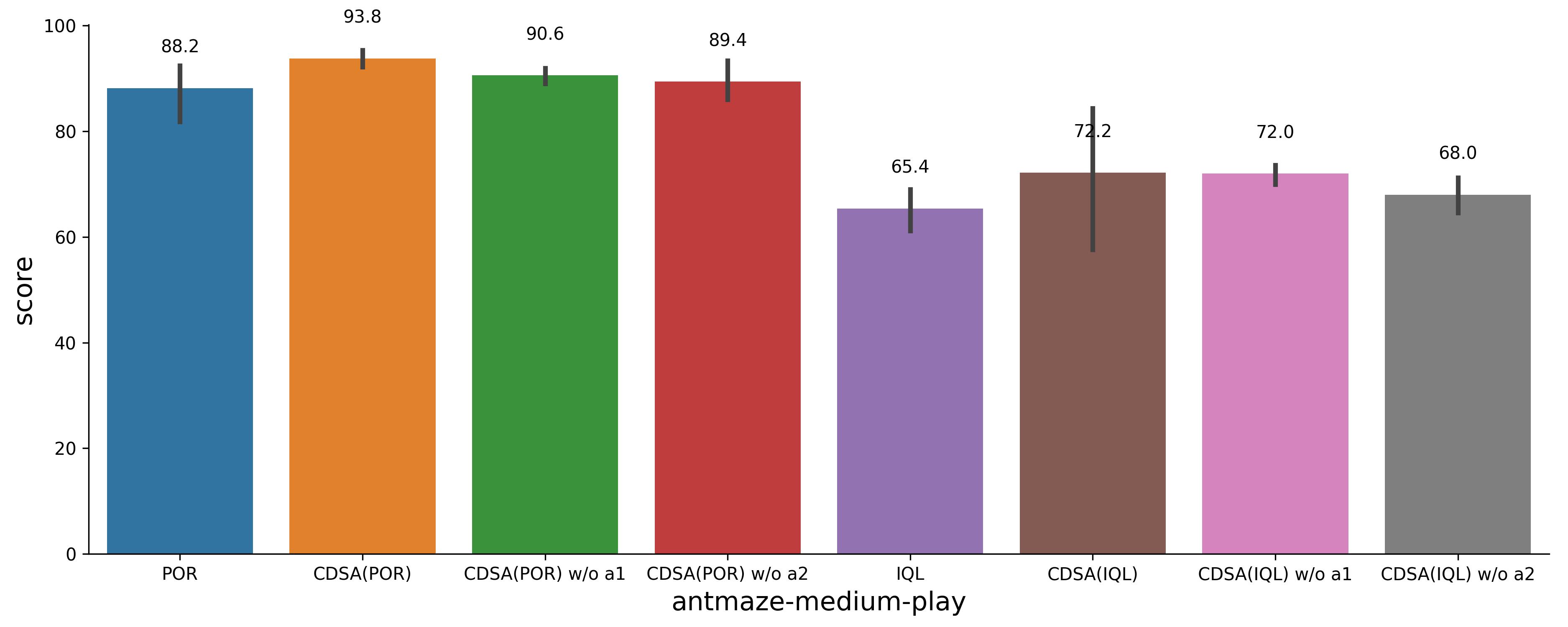}
    \end{minipage}
    }
\subfigure{
    \begin{minipage}[t]{0.3\linewidth}
    \centering
    \includegraphics[height=80pt,width=155pt]{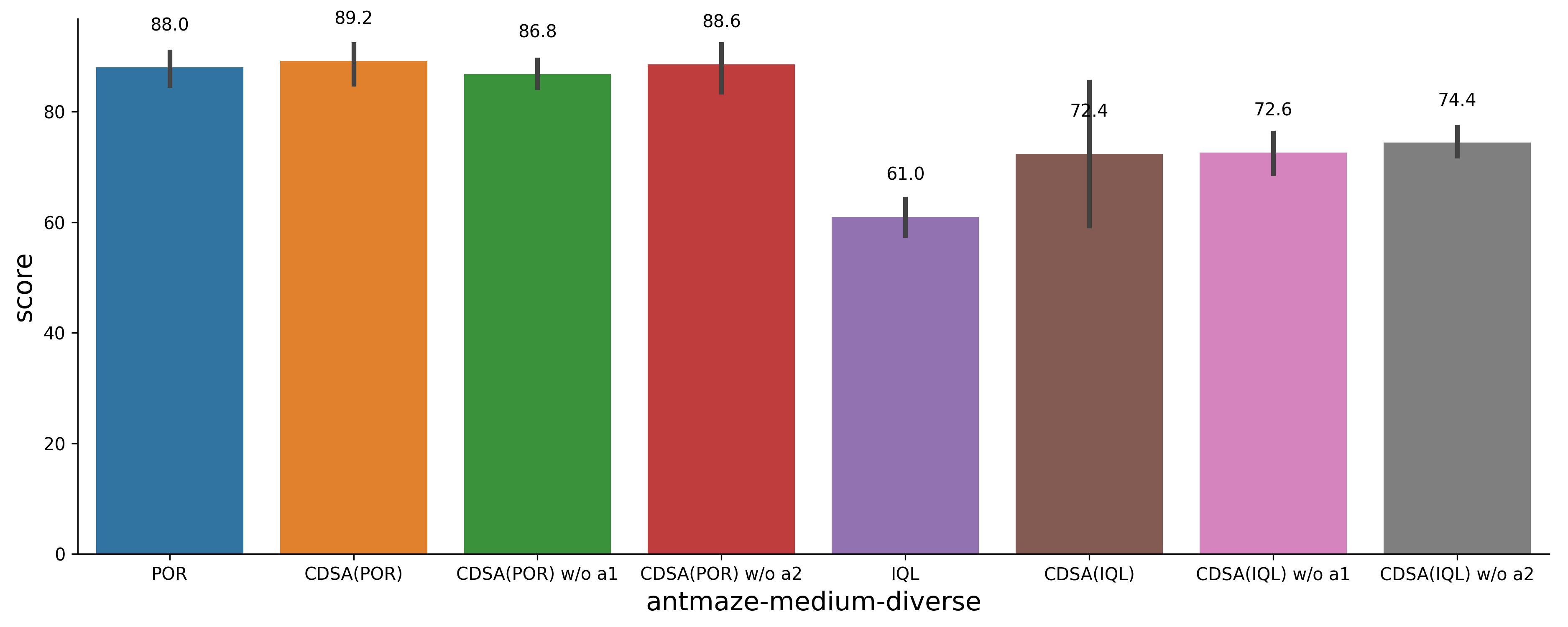}
    \end{minipage}
    }
    \subfigure{
    \begin{minipage}[t]{0.3\linewidth}
    \centering
    \includegraphics[height=80pt,width=155pt]{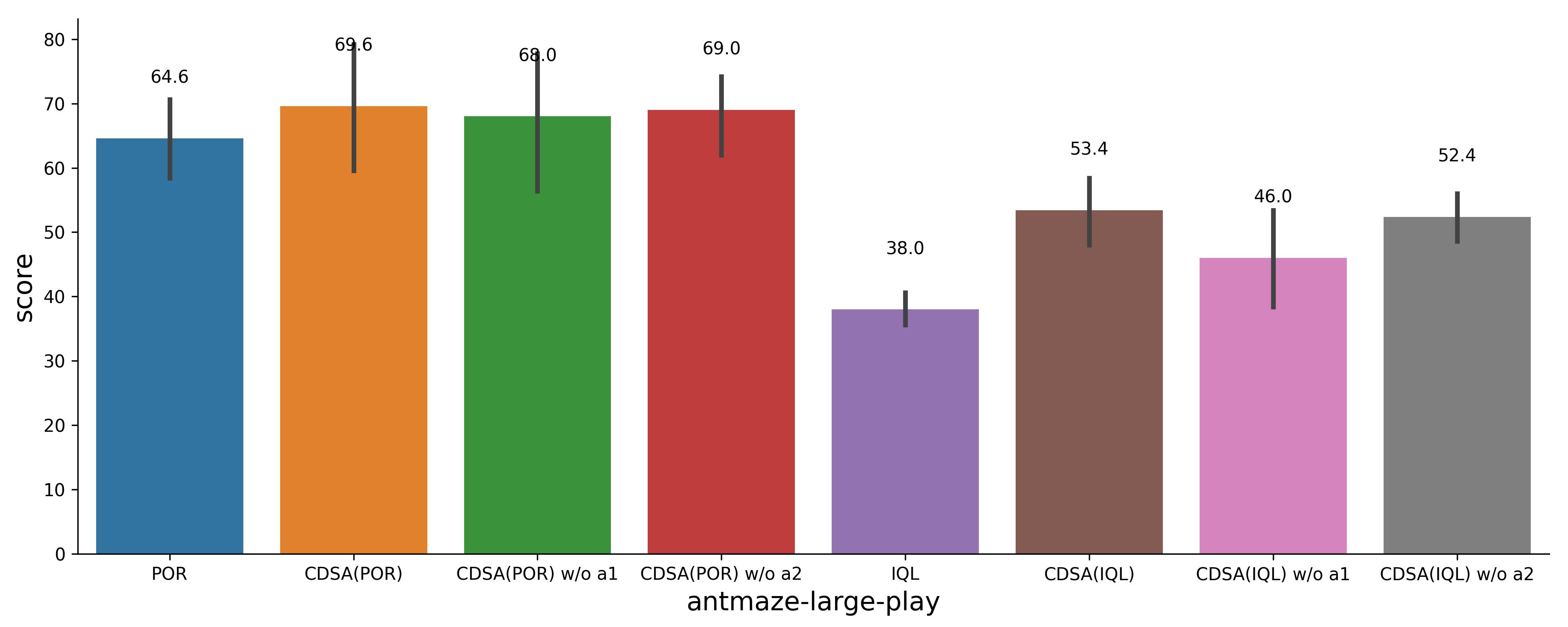}
    \end{minipage}
    }
    \subfigure{
    \begin{minipage}[t]{0.3\linewidth}
    \centering
    \includegraphics[height=80pt,width=155pt]{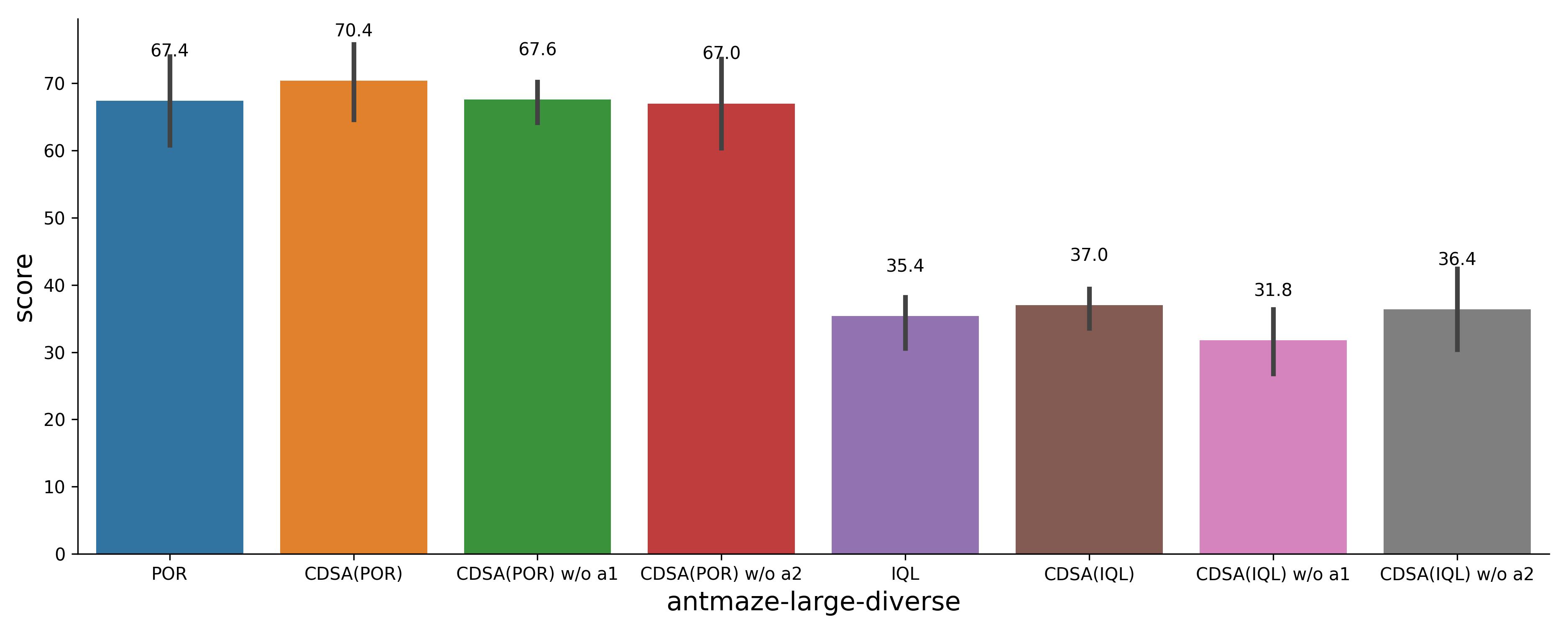}
    \end{minipage}
    }
    \subfigure{
    \begin{minipage}[t]{0.3\linewidth}
    \centering
    \includegraphics[height=80pt,width=155pt]{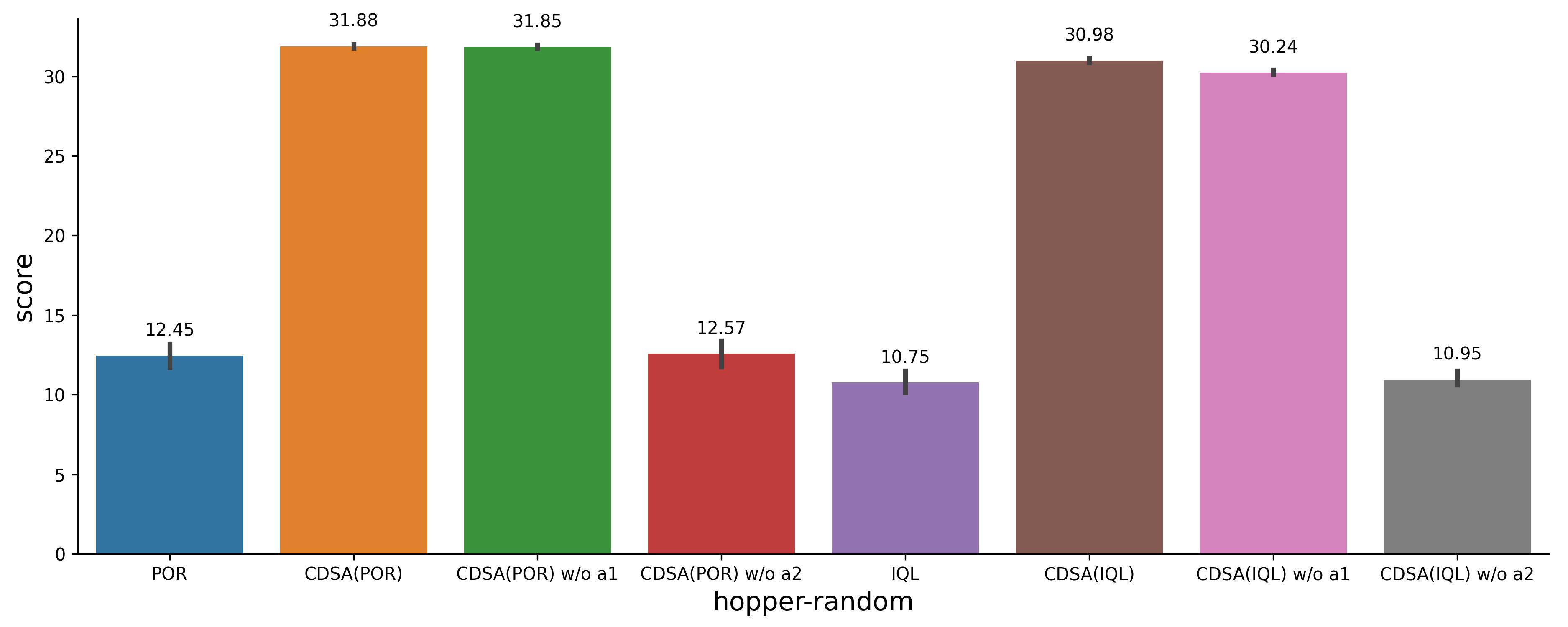}
    \end{minipage}
    }
\subfigure{
    \begin{minipage}[t]{0.3\linewidth}
    \centering
    \includegraphics[height=80pt,width=155pt]{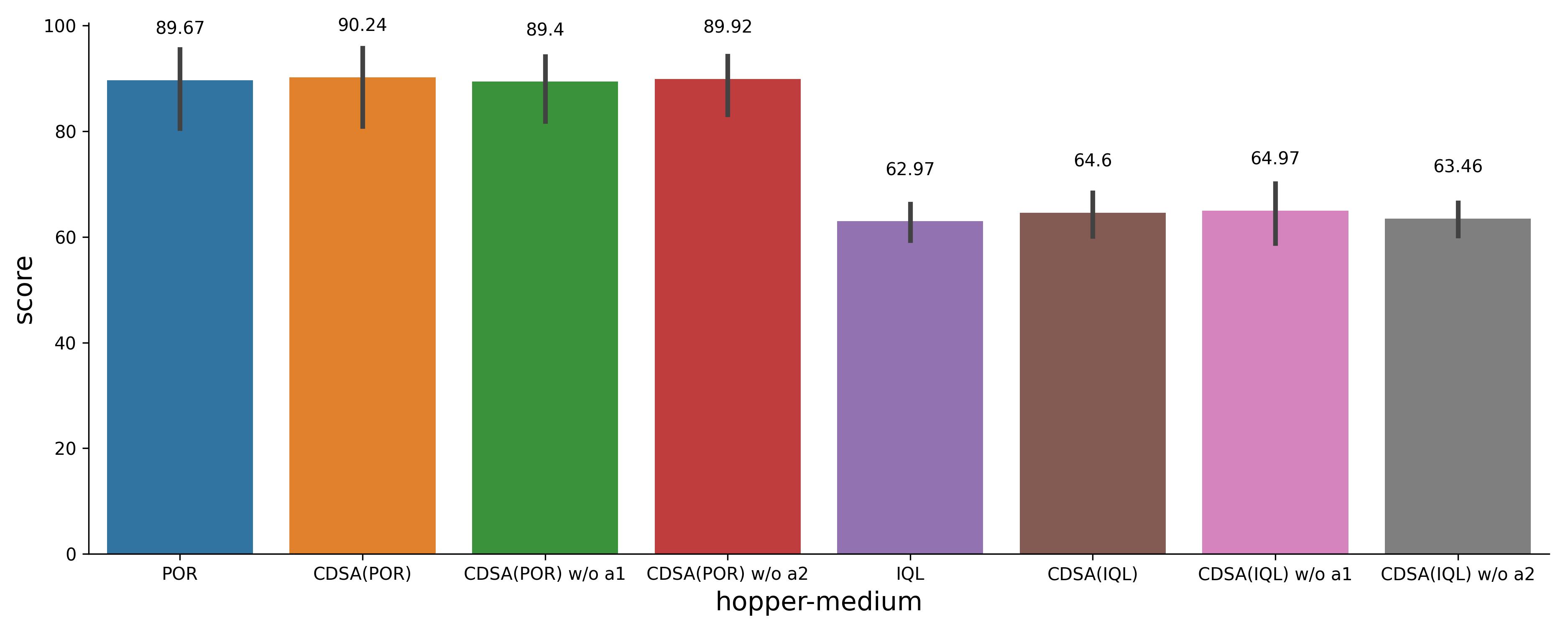}
    \end{minipage}
    }
    \subfigure{
    \begin{minipage}[t]{0.3\linewidth}
    \centering
    \includegraphics[height=80pt,width=155pt]{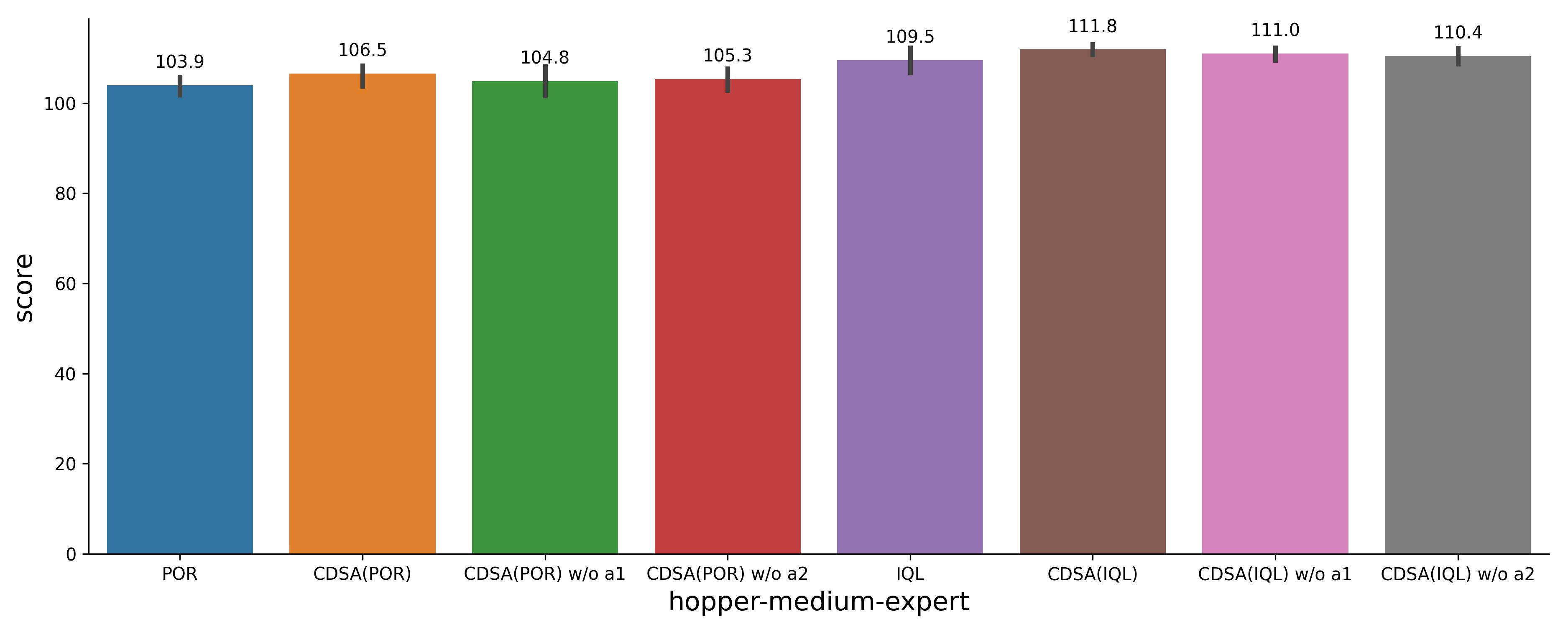}
    \end{minipage}
    }
    \subfigure{
    \begin{minipage}[t]{0.3\linewidth}
    \centering
    \includegraphics[height=80pt,width=155pt]{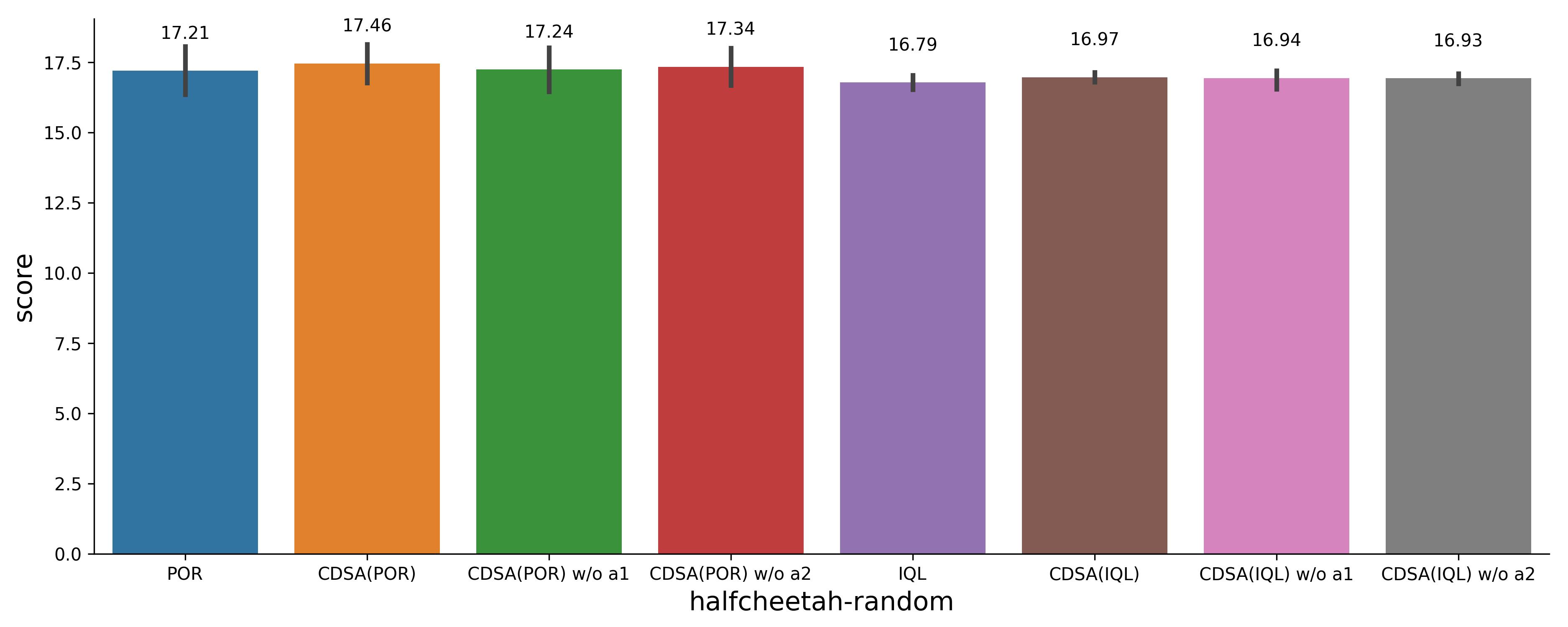}
    \end{minipage}
    }
    \subfigure{
    \begin{minipage}[t]{0.3\linewidth}
    \centering
    \includegraphics[height=80pt,width=155pt]{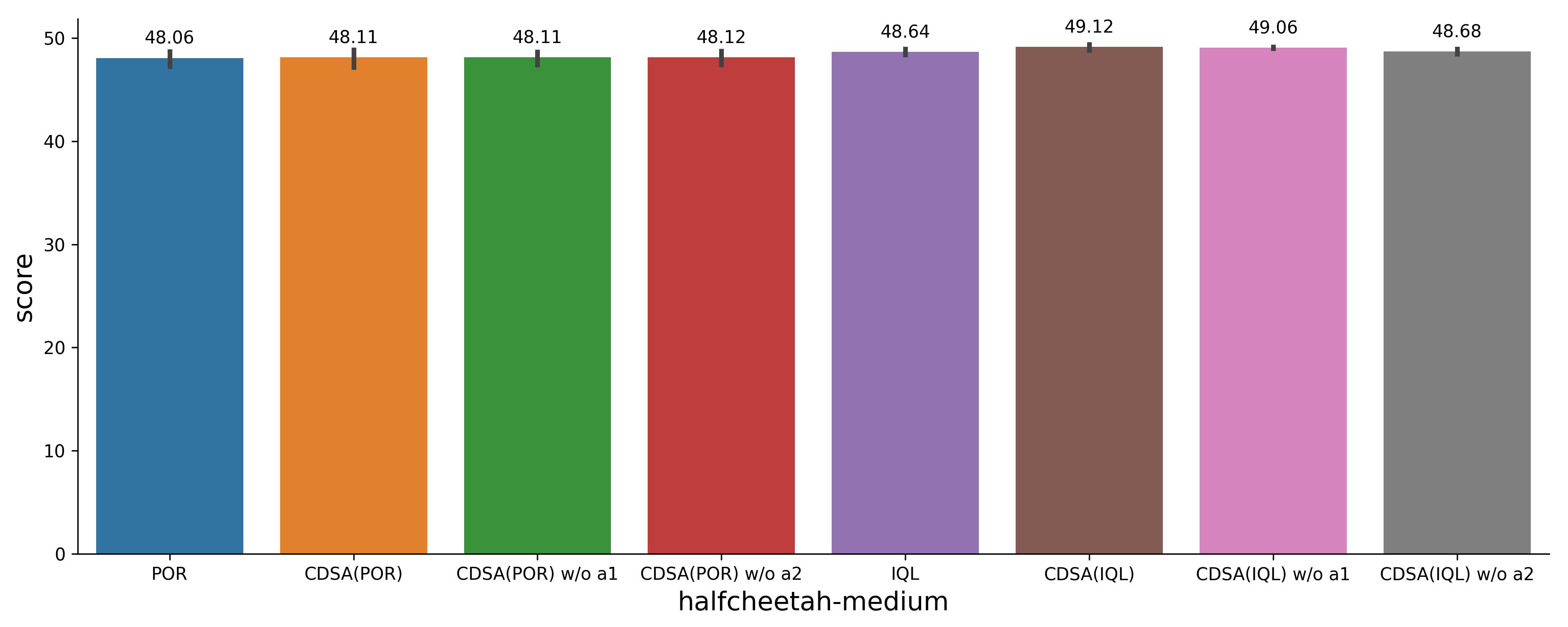}
    \end{minipage}
    }
    \subfigure{
    \begin{minipage}[t]{0.3\linewidth}
    \centering
    \includegraphics[height=80pt,width=155pt]{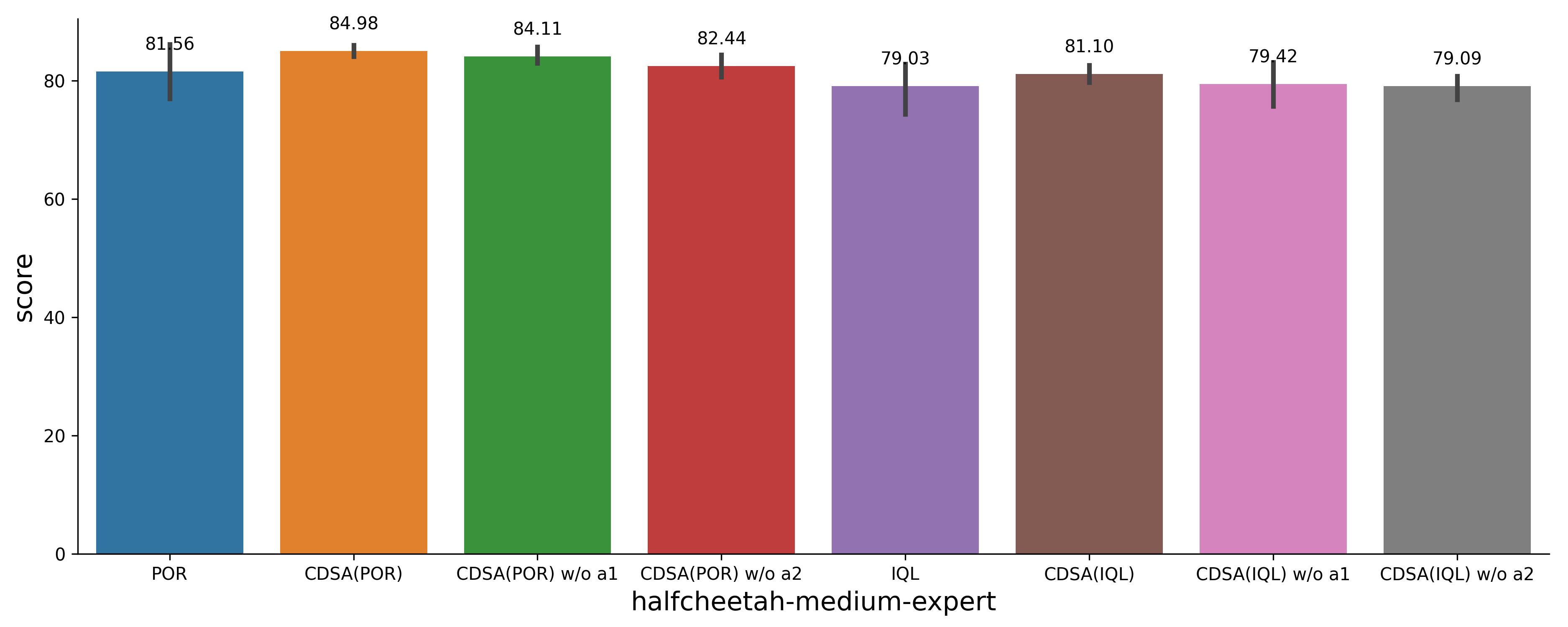}
    \end{minipage}
    }
    \subfigure{
    \begin{minipage}[t]{0.3\linewidth}
    \centering
    \includegraphics[height=80pt,width=155pt]{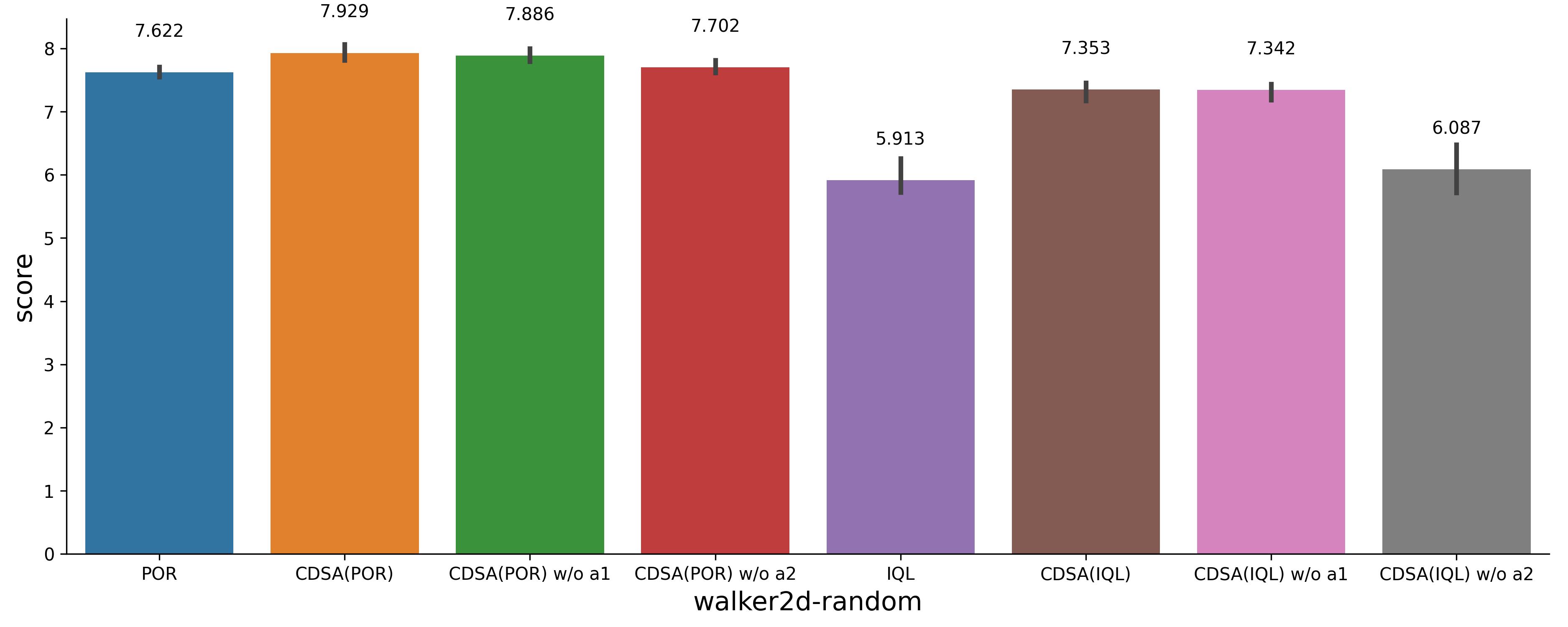}
    \end{minipage}
    }
    \subfigure{
    \begin{minipage}[t]{0.3\linewidth}
    \centering
    \includegraphics[height=80pt,width=155pt]{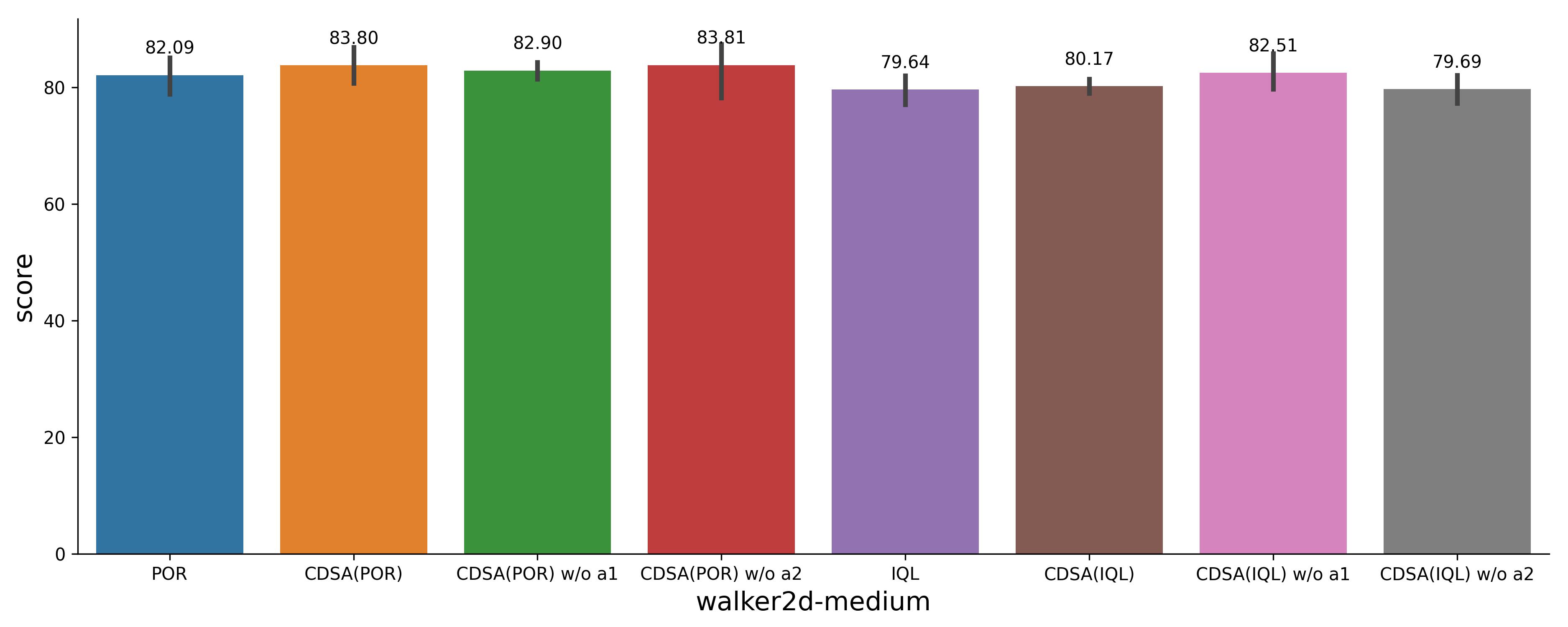}
    \end{minipage}
    }
    \subfigure{
    \begin{minipage}[t]{0.3\linewidth}
    \centering
    \includegraphics[height=80pt,width=155pt]{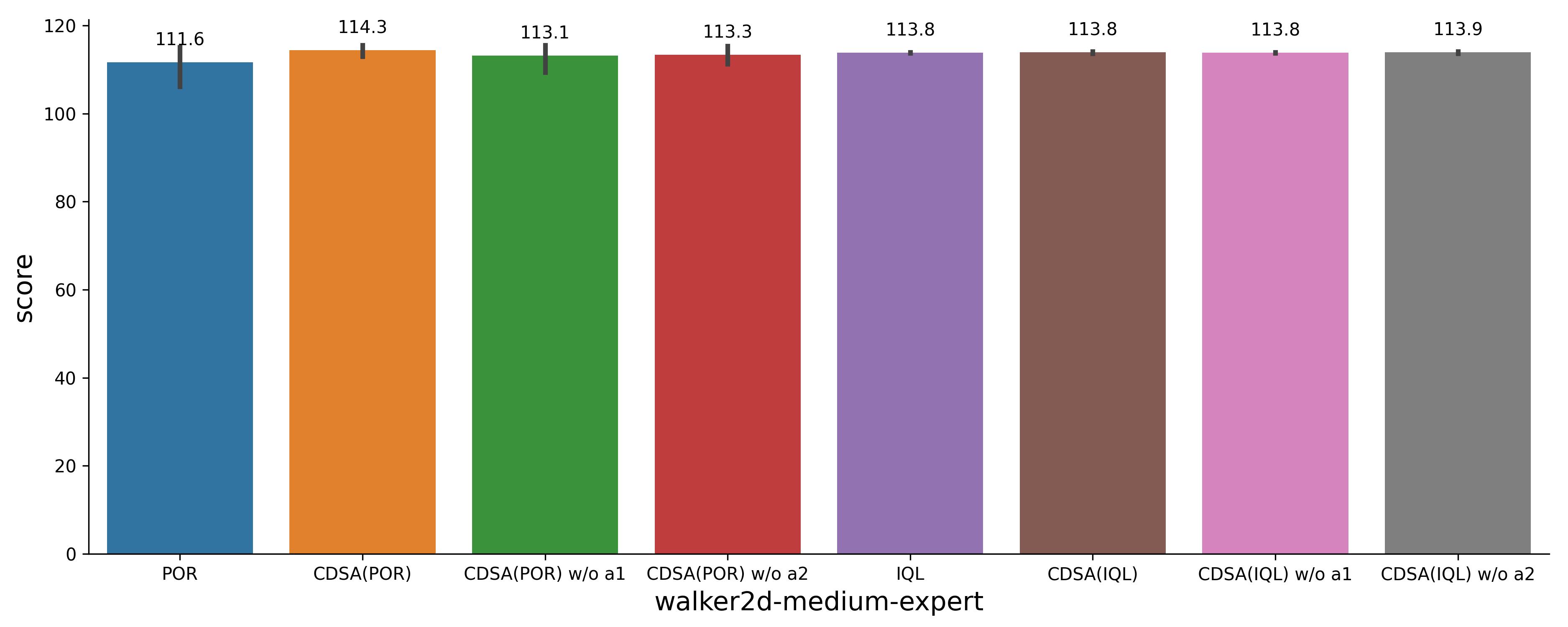}
    \end{minipage}
    }
    \caption{Results of ablation study.}
    \label{FIGURE7}
\end{figure*}

We present an ablation study to verify the effect of auxiliary actions $a_1$ and $a_2$. We denote "CDSA w/o $a_1$" only uses action $a_2$ to correct action $a_o$ and "CDSA w/o $a_2$" only uses auxiliary action $a_1$. The results are shown in Figure \ref{FIGURE7}.

We can see that the results of "CDSA w/o $a_1$" and "CDSA w/o $a_2$" are better than the baseline algorithm, which proves the effectiveness of using auxiliary action $a_1$ and $a_2$ solely. However, the performance of CDSA is better than these two algorithms and the baseline algorithm in almost every dataset, verifying it is better to use both together.

\end{document}